\documentclass[twocolumn]{article}
\usepackage{graphicx} 
\usepackage[margin=2cm]{geometry}
\usepackage{amsmath}
\usepackage{tikz}
\usepackage{subcaption}
\usepackage{enumitem}
\usepackage[T1]{fontenc}
\usepackage{authblk}
\usepackage[hidelinks]{hyperref}
\usepackage{titlesec}

\titleformat{\section}
  {\normalfont\Large\bfseries\raggedright
   \hyphenpenalty=10000\exhyphenpenalty=10000}
  {\thesection}
  {1em}
  {}

\titleformat{\subsection}
  {\normalfont\large\bfseries\raggedright
   \hyphenpenalty=10000\exhyphenpenalty=10000}
  {\thesubsection}
  {1em}
  {}

\titleformat{\subsubsection}
  {\normalfont\normalsize\bfseries\raggedright
   \hyphenpenalty=10000\exhyphenpenalty=10000}
  {\thesubsubsection}
  {1em}
  {}

\definecolor{flatorange}{RGB}{243, 156, 18}
\definecolor{flatgreen}{RGB}{39, 174, 96}
\definecolor{flatpurple}{RGB}{142, 68, 173}

\title{A Metaheuristic Optimization Framework for Discrete Optimization under Strict Time Limits}

\author[1]{Umut Çalıkyılmaz}
\author[1]{Nitin Nayak}
\author[2]{Sven Groppe}

\affil[1]{University of Lübeck, Lübeck, Germany}
\affil[2]{TU Bergakademie Freiberg, Freiberg, Germany}

\affil[ ]{\small
\texttt{umut.calikyilmaz@uni-luebeck.de},
\texttt{nitin.nayak@uni-luebeck.de}\\

\texttt{sven.groppe@informatik.tu-freiberg.de}}

\date{September 2026}

\begin{document}

\maketitle

\begin{abstract}

Real-time applications often rely on optimization approaches that can find high-quality solutions to hard problems on the order of milliseconds. Metaheuristic optimization frameworks (MOFs) are useful tools for such tasks, as they provide large sets of general-purpose search mechanisms that can return solutions under different computational budgets. However, existing work largely overlooks the available computation time as an explicit dimension of analysis. In this work, we introduce STILO, a MOF specifically designed for optimization under strict time limits. STILO integrates fine-grained configuration spaces for ant colony optimization (ACO), genetic algorithm (GA), and simulated annealing (SA), combining existing and novel operators. We performed experiments using both synthetic and benchmark instances of various discrete optimization problems. The results indicate that the proposed discrete distance calculation mechanism for SA is useful for optimization under strict time limits. They also show that the relative effectiveness of the proposed problem-independent graph structures for ACO can vary across time limits, even for the same instance characteristics. More generally, the results demonstrate that the effectiveness of algorithm families and operators depends not only on the problem type, but also on the characteristics of the instance and the available computational budget.
\vspace{2mm}
\end{abstract}

\section{Introduction}
\label{sec:introduction}

Metaheuristic algorithms are designed to find high-quality solutions to hard optimization problems in a limited time \cite{glover1989tabu}. They explore the possible solutions of a problem while evaluating them according to an objective function, and many can be applied to a variety of problems. However, the no-free-lunch theorems state that there is no single algorithm that performs best across all optimization problems \cite{wolpert2002no}. Selecting an efficient algorithm for a given instance of a problem is a non-trivial task itself, known as the algorithm selection problem \cite{rice1976algorithm}. A prominent model for solving this problem is to find a mapping from the problem space to an algorithm space based on a performance metric. For such a model, the choice of the algorithm space containing diverse methods is crucial \cite{hong2004groups}. Metaheuristic optimization frameworks (MOFs) emerged as optimization environments that offer a wide range of reusable metaheuristic methods applicable to various optimization problems \cite{parejo2012metaheuristic}. In addition to providing algorithm spaces for optimization applications, these frameworks facilitate the evaluation of not only complete algorithms but also individual algorithm structures and operators.

Real-time computing systems operate under strict time limits, and their correctness depends on the execution time as well as the quality of the results \cite{stankovic2002misconceptions}. Real-time optimization is used in many domains, including multi-agent path finding \cite{yukhnevich2026enhancing,liang2025real}, autonomous driving \cite{zhang2023optimal}, and transaction scheduling \cite{ccalikyilmaz2025optima}. In these systems, the time window for finding solutions to hard optimization problems is often on the order of milliseconds. Metaheuristic methods are well suited to such settings, since they can be terminated after any iteration and return the best solution found so far. The purpose of this study is to provide a time-aware metaheuristic optimization framework for such tasks and analyze the relative effectiveness of algorithm structures and operators across different problem types, instance characteristics, and computational budgets. Our main contributions are listed below.

\begin{itemize}[itemsep=0mm, leftmargin=4mm]
    \item We introduce STILO\footnote{\url{https://github.com/umutcalikyilmaz/STILO}}, a metaheuristic framework for optimization under strict time limits. It uses the elapsed CPU time and average iteration duration to terminate execution when another iteration is expected to exceed a user-defined CPU-time limit. The framework is also equipped with analysis tools to evaluate the solution quality obtained by a given solver configuration under various time limits.
    \item We implement a unified algorithm space consisting of ant colony optimization (ACO), genetic algorithm (GA), and simulated annealing (SA), providing a fine-grained configuration space of various hyperparameters and modular operators for each algorithm family.
    \item We contribute two general-purpose graph structures for ACO that can be applied without problem-specific graph design. We also integrate a discrete distance calculation mechanism into SA and propose various distance calculation operators inspired by continuous SA applications.
    \item We conduct an empirical study to evaluate the performance of algorithm families and operators across various problem types, instance characteristics, and time limits on the order of milliseconds.
\end{itemize}

The results of the analysis provide important insights into time-limited optimization with metaheuristic methods. The most general result is that the relative effectiveness of the tested algorithm families and operators depends on the time limit, problem type, and instance characteristics. Each tested algorithm family achieved the best solution quality for at least some time limits and instance classes, while almost all tested operators were used in the best-performing configurations for at least some settings. Another noteworthy result is that most of the best GA and SA configurations used combinations of operators for the majority of problem settings, indicating the usefulness of operator mixing for some steps of these algorithm families. The results also show that, although the graph structure for ACO depends on the structure of the target solution space, the available computational budget can affect this choice for some instances. Finally, the results strongly support the usefulness of the discrete distance calculation mechanism for SA in time-limited optimization, as Uniform, Gaussian, or Cauchy distance calculation operators were selected by the best-performing configurations substantially more often than the default Constant distance operator.

The remainder of the paper is organized as follows. Section~\ref{sec:related} reviews related work on metaheuristic optimization frameworks. Section~\ref{sec:problem} describes the problem setting targeted by the proposed algorithm space and briefly introduces the combinatorial problems used in the experiments. Section~\ref{sec:metaheuristic} introduces the three metaheuristic algorithm families considered in this study and presents the configuration space created for each family. Section~\ref{sec:experimental} presents the results of the experiments conducted on synthetic and benchmark instances and discusses their implications. Finally, Section~\ref{sec:conclusion} summarizes the main conclusions and outlines directions for future work.

\section{Related Work}
\label{sec:related}

Metaheuristic optimization frameworks (MOFs) are software tools that offer a set of reusable metaheuristic algorithms to solve optimization problems and conduct experiments \cite{parejo2012metaheuristic}. These frameworks streamline the implementation process by providing diverse sets of algorithms, and some also contain additional tools for performance monitoring, algorithm selection, and possibly other tasks. They separate problem-specific solution representations and metaheuristic components, and most of these frameworks provide general-purpose algorithms that are applicable to a broad range of problems, although problem-specific frameworks also exist, such as SATenstein for the Boolean satisfiability problem \cite{khudabukhsh2016satenstein}.

MOFs can be roughly divided into three groups according to the metaheuristic spaces they offer. The frameworks in the first group provide modular operators that can be used to configure a single algorithm family. For example, \cite{van2017algorithm} offers 4,608 variations of the covariance matrix adaptation evolution strategy by turning 11 modules on or off. Similarly, \cite{lopez2012automatic} and \cite{campelo2020moeadr} provide various components to configure multi-objective ACO and multi-objective evolutionary algorithms, respectively. MOFs that use multiple algorithm families form the second group. FOM \cite{parejo2003fom} and Opt4J \cite{lukasiewycz2011opt4j} implement several well-known metaheuristic structures and multiple interchangeable operators for discrete optimization. Another example is jMetal \cite{durillo2011jmetal}, which is built around four metaheuristic algorithm families for multi-objective optimization. Long-running projects such as ParadisEO \cite{dreo2021Paradiseo}, HeuristicLab \cite{wagner2005heuristiclab}, and ECJ \cite{scott2019ecj} have been extended over the years, and their current versions support numerous algorithm families and operators. Frameworks such as AutoOpt \cite{zhao2025autoopt} and METAFOR \cite{camacho2025metafor} are in the third group, and are often called automated algorithm generation frameworks. These frameworks can create different algorithm structures by combining operators and computational primitives, instead of using fixed structures imposed by existing algorithm families. The current version of ParadisEO also contains algorithm generation \cite{dreo2021Paradiseo}.

STILO belongs to the second group, as it uses three algorithm families (ACO, GA, SA) and various operators to configure them. The distinguishing feature of STILO is its explicit treatment of strict time limits and the tools it provides to assess the performance of different structures and operators under varying time limits. Some previous studies have also considered computational budget as an important factor for MOFs. For example, in the experiments presented in \cite{durillo2011jmetal}, the execution of each algorithm is limited to a constant number of cost function evaluations for a fair comparison. However, the number of evaluations itself is not the sole factor contributing to execution time. Different operators and structures may incur different amounts of computational overhead, and for optimization under strict time limits, these should be taken into consideration. For this purpose, we propose using CPU time as a measure of computational budget. A similar approach is adopted in \cite{camacho2025metafor} by setting a wall-clock termination condition, but only a single time limit is used. In this study, we analyze the performance of numerous configurations under different time limits ranging from 50 ms to 500 ms and examine the effectiveness of different algorithm families and operators under these limits.

\section{Problem Setting and Benchmark Problems}
\label{sec:problem}

STILO serves as a general-purpose optimization environment for a broad range of discrete single-objective optimization problems. To achieve this, each algorithm is considered a black-box approach, and the target problems are abstracted through two components. The first component is the structure of the solution space. The framework supports problems whose solutions can be represented as strings of integer values. The structure of the corresponding solution space is defined by three parameters: the length of the integer string $sl$, the number of states that each integer can assume $sn$, and a Boolean parameter $u$ indicating whether the integer values in a string must be unique. As shown in Section~\ref{sec:metaheuristic}, these parameters are used to determine the behavior of the operators and to identify which operators are applicable to a given problem. The second component is a cost function $C(x)$, which returns the cost of the solution $x$. In addition to these general problem representation components, ACO may optionally use problem-specific information.

In the following, we introduce the four benchmark problems used to evaluate the configurations generated by our framework and show how their instances are represented using the given problem representation components.

\subsection{Identical Machines Scheduling Problem (IMSP)}

IMSP is the problem of assigning a set of jobs to multiple identical machines in a way that minimizes the maximum machine completion time, which is called the makespan \cite{graham1969bounds}. Also known as the multi-way partitioning problem, IMSP is NP-hard \cite{korf2009multi}.
An instance of IMSP is defined by the number of jobs $n$, the number of machines $m$, and a $1\times n$ matrix $\boldsymbol{L}$ that contains the lengths of the jobs. A solution to the problem is an assignment of each of the $n$ jobs to one of the $m$ machines, since there is no constraint that makes the order of jobs significant. The structure of the search space for this instance is defined as $sl=n$, $sn=m$, and $u=false$. The $i^{th}$ integer in a solution string indicates the machine to which job $i$ is assigned. The cost function returns the maximum of the total lengths of the jobs assigned to the machines.

\subsection{Max-Cut Problem (MCP)}

MCP is the problem of dividing the vertices of a given undirected graph into two groups such that the total weight of the edges between these groups is maximized \cite{commander2009maximum}, and it is NP-hard \cite{ageev2014complexity}. In the rest of the paper, we use MCP to refer to the unweighted version of the problem, where the edge weights can only be $0$ or $1$, and weighted MCP (WMCP) to refer to the general weighted version. An instance of MCP or WMCP is defined by the number of vertices $n$ and the weight matrix $\boldsymbol{E}$, which is an $n \times n$ symmetric matrix containing binary values for MCP and real-valued weights for WMCP. Possible solutions to an MCP or WMCP instance can be represented by a string of binary values, where the $i^{th}$ value indicates the assignment of the $i^{th}$ vertex, with $0$ and $1$ representing the first and second groups, respectively. Thus, the structure parameters of the solution space are $sl=n$, $sn=2$, and $u=false$. MCP and WMCP are maximization problems, so their cost function is defined as $C(x)=M-W(x)$, where $M$ is the total positive edge weight and $W(x)$ is the sum of the weights of the edges between the two groups determined by the solution $x$. Although the algorithms operate on $C(x)$, we report the original cut weight $W(x)$ in the experimental results presented in Section~\ref{sec:experimental} for better interpretation. 

\subsection{Transaction Scheduling Problem (TxnSP)}

TxnSP is an NP-hard job scheduling problem in an identical parallel machine setting with the objective of minimizing the makespan, where some pairs of jobs conflict \cite{ccalikyilmaz2023opportunities,ccalikyilmaz2025optima}. Two conflicting jobs are forbidden to be processed simultaneously, so in TxnSP, changing the order of the jobs can result in effectively different schedules, unlike IMSP. A TxnSP instance is defined by the number of jobs $n$, the number of machines $m$, the length matrix $\boldsymbol{L}$, and the conflict matrix $\boldsymbol{Q}$. $\boldsymbol{L}$ is a $1\times n$ matrix of continuous values containing the lengths of the jobs. $\boldsymbol{Q}$ is an $n \times n$ symmetric matrix of binary values, where $Q_{ij}=1$ if jobs $i$ and $j$ conflict, and $Q_{ij}=0$ otherwise. Possible solutions to a TxnSP instance can be represented by a permutation of jobs. Thus, a solution is represented by a string of $n$ unique integer values, and the parameters of the solution space are $sl=n$, $sn=n$, and $u=true$. The cost function $C(x)$ decodes the permutation $x$ to create a schedule and returns its makespan.

\subsection{Traveling Salesperson Problem (TSP)}

TSP is the problem of finding the Hamiltonian cycle with the minimum length for a given weighted graph \cite{gavish1978travelling}. In this paper, we only consider the symmetric TSP variant, which is NP-hard \cite{karp2010reducibility}. The defining properties of a TSP instance are the number of cities $n$ and an $n \times n$ symmetric matrix $\boldsymbol{D}$, which contains the distances between cities. Each solution is a sequence of visited cities. Since each tour is a closed loop, one city can be fixed as the starting city, and each tour can be represented by a string of $n-1$ unique integers corresponding to the remaining cities. The structure of this problem instance is defined as $sl=n-1$, $sn=n-1$, and $u=true$. The cost function $C(x)$ constructs the tour represented by $x$ and returns its total distance.

\section{Metaheuristic Configuration Spaces}
\label{sec:metaheuristic}

The optimization environment of STILO is built around three algorithm families: ACO, GA, and SA. For each family, we integrated a modular algorithm structure with a fine-grained configuration space consisting of tunable hyperparameters and interchangeable operators for different calculation steps. The purpose of these configuration spaces is to provide a diverse set of choices for optimizing a wide range of problems under various time constraints. Some of the operators are specifically designed for $u=true$ or $u=false$ solution spaces, while most are applicable to both.

In the following, we briefly introduce the algorithm families, present their implemented structures, and list the hyperparameters and operators included in their configuration spaces.

\subsection{Ant Colony Optimization}
\label{ssec:aco}

\begin{figure*}[t]
    \centering
    \includegraphics[width=0.7\linewidth]{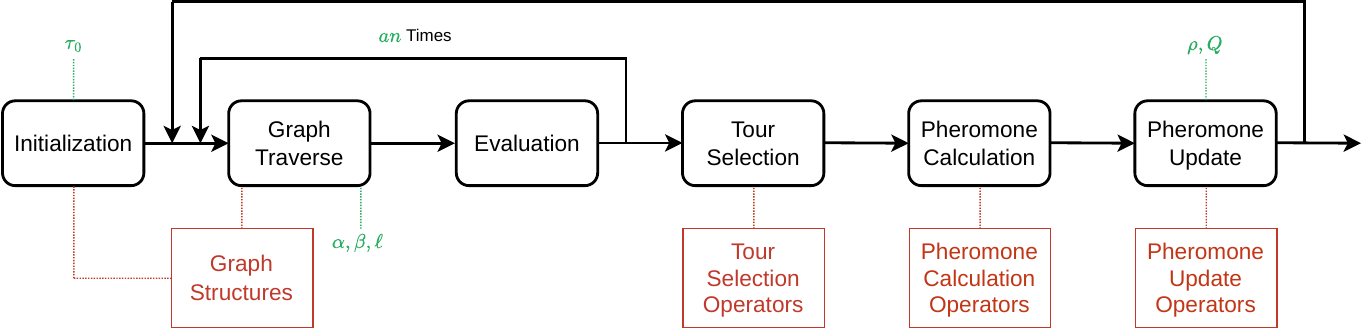}
    \caption{The structure of the ant colony optimization implementation. }
    \label{fig:aco}
\end{figure*}

Ant colony optimization (ACO) is a metaheuristic inspired by the foraging behavior of ant colonies \cite{dorigo1997ant}. It uses simulated ants that traverse a graph and update the pheromone levels on the edges to drive the search toward high-quality solutions. Figure~\ref{fig:aco} shows the structure of the ACO implementation in our framework, indicating the hyperparameters (green) and the operator sets (red) that form the configuration space. In the following, we explain the components of the ACO configuration space and their functions in the implemented structure.

\subsubsection{General Hyperparameters}

The initial pheromone ($\tau_0$) specifies the amount of pheromone assigned to each edge of the graph during initialization. The number of ants ($an$) determines how many times the graph is traversed in each iteration. The evaporation parameter ($\rho$) and the pheromone constant ($Q$) influence the pheromone update process. The pheromone influence ($\alpha$) and the heuristic influence ($\beta$) are used in the stochastic edge selection process during traversal. The local decay parameter ($\ell$) is used for local pheromone updates during traversal to increase the probability of discovering new solutions, as proposed in \cite{dorigo1997ant}. If $\ell=0$, the local update is not applied.

\subsubsection{Graph Structures}

At each iteration, each ant traverses the graph once to construct a solution. The path of an ant is determined stochastically based on the current pheromone levels and problem-specific heuristic information. When ant $k$ is at vertex $i$, the probability of moving to edge $ij$ is calculated as

\begin{equation}
    p_{ij}^k=\frac{(\tau_{ij})^\alpha(\eta_{ij})^\beta}{\sum_{l\in \textbf{A}(i)} (\tau_{il})^\alpha(\eta_{il})^\beta},
    \label{eq:edgeprob}
\end{equation}
where $\textbf{A}(i)$ is the set of vertices connected to $i$, and $\tau_{ij}$ and $\eta_{ij}$ are the pheromone level and the heuristic likelihood of edge $ij$. The heuristic likelihoods are problem-specific, and their definitions for the problems considered in this study are provided in the STILO implementation. Each traversal creates a solution $x$, which is evaluated by the cost function $C(x)$.

We implemented two general-purpose graph structures that can be used for any instance in the target problem setting. These structures are independent of problem-specific details and are defined by the solution-space parameters introduced in Section~\ref{sec:problem}. The two graph structures are described below.

\begin{figure}[t]
    \centering
    \begin{subfigure}{0.48\textwidth}
        \centering
        \begin{tikzpicture}
        \draw (0,0) circle (8pt);
        \draw (2,1) circle (8pt);
        \draw (2,-1) circle (8pt);
        \draw (4,-0.6) circle (8pt);
        \draw (4,0.6) circle (8pt);

        \node at (0,0) {s};
        \node at (2, 1) {0};
        \node at (2, -1) {1};
        \node at (4, 0.6) {2};
        \node at (4, -0.6) {3};

        \draw[->, color=gray] (0.3,0.2) -- (1.65,0.9);
        \draw[->, color=gray] (0.3,-0.2) -- (1.65,-0.9);
        \draw[->, line width=1] (0.35,0.1) -- (3.65,0.55);
        \draw[->, color=gray] (0.35,-0.1) -- (3.65,-0.55);

        \draw[->, color=gray] (1.94,0.65) -- (1.94,-0.65);
        \draw[->, line width=1] (2.06,-0.65) -- (2.06,0.65);

        \draw[->, color=gray] (3.94,0.25) -- (3.94,-0.25);
        \draw[->, color=gray] (4.06,-0.25) -- (4.06,0.25);

        \draw[->, color=gray] (2.35,0.96) -- (3.65,0.66);
        \draw[->, color=gray] (3.65,0.8) -- (2.35,1.1);

        \draw[->, color=gray] (2.35,-0.96) -- (3.65,-0.66);
        \draw[->, color=gray] (3.65,-0.8) -- (2.35,-1.1);

        \draw[->, line width=1] (2.3,0.85) -- (3.8,-0.3);
        \draw[->, color=gray] (3.73,-0.4) -- (2.23,0.76);

        \draw[->, color=gray] (2.3,-0.85) -- (3.8,0.3);
        \draw[->, line width=1] (3.73,0.4) -- (2.23,-0.76);

        \draw[->, color=gray] (2.25,1.28) arc (-30:210:0.3);
        \draw[->, color=gray] (2.25,-1.28) arc (30:-210:0.3);
        \draw[->, color=gray] (4.38,-0.68) arc (60:-180:0.3);
        \draw[->, color=gray] (4.38,0.68) arc (-60:180:0.3);
                
    \end{tikzpicture}
    \caption{An example tour on the Single-Stage graph structure}
    \label{fig:singlestage}
    \end{subfigure}

    \vspace{2mm}
    \begin{subfigure}{0.48\textwidth}
        \centering
        \begin{tikzpicture}
        \draw (0,0) circle (8pt);
        \node at (0,0) {s};

        \draw (1.1,0.5) circle (8pt);
        \node at (1.1,0.5) {0};

        \draw (1.1,-0.5) circle (8pt);
        \node at (1.1,-0.5) {1};

        \draw (2.2,0.5) circle (8pt);
        \node at (2.2,0.5) {0};

        \draw (2.2,-0.5) circle (8pt);
        \node at (2.2,-0.5) {1};

        \draw (3.3,0.5) circle (8pt);
        \node at (3.3,0.5) {0};

        \draw (3.3,-0.5) circle (8pt);
        \node at (3.3,-0.5) {1};

        \draw (4.4,0.5) circle (8pt);
        \node at (4.4,0.5) {0};

        \draw (4.4,-0.5) circle (8pt);
        \node at (4.4,-0.5) {1};

        \draw[->, color=gray] (0.35,0.1) -- (0.75,0.4);
        \draw[->, line width=1] (0.35,-0.1) -- (0.75,-0.4);

        \draw[->, color=gray] (1.45,0.6) -- (1.85,0.6);
        \draw[->, color=gray] (1.45,0.4) -- (1.85,-0.4);
        \draw[->, color=gray] (1.45,-0.4) -- (1.85,0.4);
        \draw[->, line width=1] (1.45,-0.6) -- (1.85,-0.6);

        \draw[->, color=gray] (2.55,0.6) -- (2.95,0.6);
        \draw[->, color=gray] (2.55,0.4) -- (2.95,-0.4);
        \draw[->, line width=1] (2.55,-0.4) -- (2.95,0.4);
        \draw[->, color=gray] (2.55,-0.6) -- (2.95,-0.6);

        \draw[->, color=gray] (3.65,0.6) -- (4.05,0.6);
        \draw[->, line width=1] (3.65,0.4) -- (4.05,-0.4);
        \draw[->, color=gray] (3.65,-0.4) -- (4.05,0.4);
        \draw[->, color=gray] (3.65,-0.6) -- (4.05,-0.6);
    \end{tikzpicture}
    \caption{An example tour on the Multi-Stage graph structure}
    \label{fig:multistage}
    \end{subfigure}
    \caption{Graph structures in the ant colony optimization implementation}
    \label{fig:graph}
\end{figure}
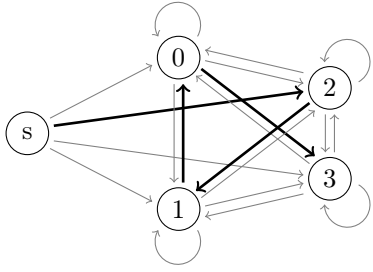
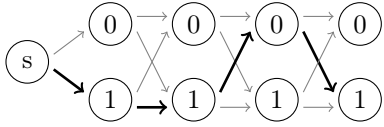

\begin{enumerate}[itemsep=1mm, leftmargin=0mm, label={}]
    \item (1) \textbf{Single-Stage Graph:} A Single-Stage graph has $sn+1$ nodes: one source node and $sn$ nodes, each of which represents an integer value in the solution string. Starting from the source node, an ant makes $sl$ moves, and the node visited in the $i^{th}$ move determines the value of the $i^{th}$ integer of the constructed solution. If $u=true$, previously visited nodes are forbidden for the rest of the same tour. The Single-Stage graph structure provides a compact graph with low computational cost, but it does not store positional information. An example tour on a Single-Stage graph is given in Figure~\ref{fig:singlestage}.

    \item (2) \textbf{Multi-Stage Graph:} A Multi-Stage graph consists of a source node and $sl$ layers, each having $sn$ nodes. An ant visits one node in each layer. The index of the visited node in the $i^{th}$ layer gives the $i^{th}$ integer of the solution string. For problems with $u=true$, nodes that represent already used values are forbidden for the rest of the tour. A Multi-Stage graph stores positional information in addition to relations between different values, but it incurs higher computational overhead for problems with large $sn$, since it maintains and updates pheromone values on many edges. An example tour on a Multi-Stage graph is shown in Figure~\ref{fig:multistage}.
\end{enumerate}

\subsubsection{Tour Selection Operators}

After traversal, some resulting solutions are selected to be used for pheromone deposition. The configuration space contains three tour selection operators for this purpose, which are listed below.

\begin{itemize}[itemsep=1mm, leftmargin=0mm, label={}]
    \item (1) \textbf{Every Ant:} This is the operator that is used in the original ant system paper \cite{colorni1994ant}. All tours in each iteration are used for pheromone update.
    \item (2) \textbf{Iteration Best:} A constant number of the best tours from the current iteration are selected. The number of selected tours ($lb$) is a hyperparameter. This operator is offered by the max-min ant system variant \cite{stutzle2000max}.
    \item (3) \textbf{Global Best:} A constant number of the best solutions found over all iterations is kept. At each iteration, the list is updated and pheromone is deposited on the edges visited in the stored tours. The number of saved tours ($gb$) is a hyperparameter. This operator is used in the elitist ant system variant \cite{negulescu2008elitist}.
    
\end{itemize}

\subsubsection{Pheromone Calculation Operators}

For each selected tour, the amount of pheromone to be deposited is calculated using one of the two operators in the configuration space, which are listed below.

\begin{itemize}[itemsep=1mm, leftmargin=0mm, label={}]
    \item (1) \textbf{Cost-Based Calculation:} For each selected tour, the amount of pheromone deposit is calculated as $\Delta\tau =Q/C(x)$. This operator is proposed in the original ant system and is used in most variants \cite{colorni1994ant}.
    
    \item (2) \textbf{Rank-Based Calculation:} The selected tours are ranked according to their costs and the amount of pheromone deposit for each tour is calculated as $\Delta\tau=Q(\omega-r)/\omega$, where $r$ is the rank of the tour and $\omega$ is the maximum possible rank. This operator follows the idea of using ranks for pheromone calculation, as in the Rank-Based ant system \cite{bullnheimer1999new}.
\end{itemize}

\subsubsection{Pheromone Update Operators} The last operation of each iteration is to update the pheromone levels on the edges of the graph. For this operation, one of the two operators can be selected, which are explained below.

\begin{itemize}[itemsep=1mm, leftmargin=0mm, label={}]
    \item (1) \textbf{Classical Pheromone Update:} First, the pheromone level on each edge is evaporated as $\tau\rightarrow(1-\rho)\tau$. Then, pheromone depositions $\Delta\tau$ are added to the edges that are included in the selected tours. This operator is used in the original ant system and in most other variants \cite{colorni1994ant}.
    \item (2) \textbf{Max-Min Pheromone Update:} This operator applies the same evaporation and deposition steps as the Classical pheromone update. Then it checks whether the pheromone level on each edge is in the range $[\tau_{min},\tau_{max}]$. If the current pheromone level is outside this range, its value is set to $\tau_{min}$ or $\tau_{max}$. This operator is proposed for the max-min ant system \cite{stutzle2000max}. We implemented the methods given in that study to calculate the values of $\tau_{min}$ and $\tau_{max}$ in each iteration.
\end{itemize}

\subsection{Genetic Algorithm}

\begin{figure*}[t]
    \centering
    \includegraphics[width=0.7\linewidth]{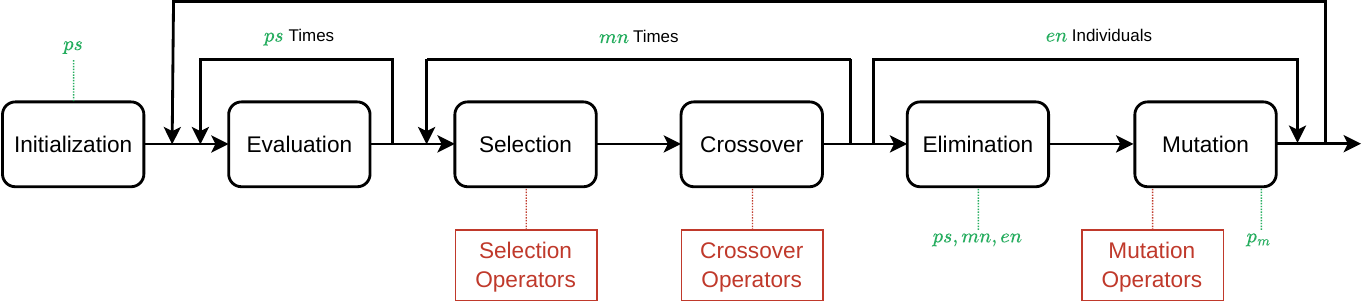}
    \caption{The structure of the genetic algorithm implementation}
    \label{fig:ga}
\end{figure*}

Inspired by genetic evolution, the genetic algorithm (GA) evolves a population of candidate solutions to obtain high-quality solutions \cite{holland1992genetic}. Figure~\ref{fig:ga} shows the structure used for GA in our framework, indicating the hyperparameters (green) and operator sets (red) that form its configuration space. In the following, we explain the roles of these components in detail.

\subsubsection{General Hyperparameters} 

The population size ($ps$) determines the number of individuals in the population. During initialization, $ps$ random solutions are created, each of which is evaluated at the beginning of each iteration. The mating-pair count ($mn$) specifies the number of parent pairs used for the crossover in each generation. The elite count ($en$) is used for elitist selection, which is offered in \cite{goldberg1991comparative}. In each generation, the best $en$ individuals are passed on to the next generation without elimination or mutation. The number of eliminated solutions is determined as $ps-2mn-en$ to keep the population size fixed over generations.

\subsubsection{Selection Operators} In each generation, a selection operator chooses pairs of solutions to participate in crossover and create offspring solutions. The selection operators offered in the configuration space are listed below.

\begin{itemize}[itemsep=1mm, leftmargin=0mm, label={}]
\label{par:selection}
    \item (1) \textbf{Steady-State Selection:} Steady-state selection is a deterministic process, where individuals are ordered by their costs and the best individuals are selected for crossover \cite{durillo2009effect}.
    
    \item (2) \textbf{Roulette Wheel Selection:} Each solution is assigned a probability value depending on its cost, and two solutions are randomly selected according to these probabilities \cite{yu2016improved}.
    
    \item (3) \textbf{Rank Selection:} Rank selection is a modified version of Roulette wheel selection, where solutions are ranked according to their costs and the rank values are used to calculate the selection probabilities \cite{abdulal2009improved}. 
    
    \item (4) \textbf{Tournament Selection:} From the population, $t$ individuals are selected randomly. The best two individuals in the sample are used for crossover \cite{goldberg1991comparative}.
\end{itemize}

\subsubsection{Crossover Operators} Crossover operators are used to create two offspring solutions from two parents. Our implementation allows combining multiple crossover operators by setting the propensity of each operator for a configuration. Before each crossover operation, one of the operators is randomly selected according to these propensities. The crossover operators integrated into the framework are shown below.

\begin{itemize}[itemsep=1mm, leftmargin=0mm, label={}]
    \item (1) \textbf{K-Point Crossover:} Both parent chromosomes are cut at $k$ random points, and the pieces are swapped between them to generate two offspring solutions \cite{de1992formal}. This operator is not applicable when $u=true$.
    
    \item (2) \textbf{Uniform Crossover:} When creating the first offspring, each gene is randomly selected from one of the parents, with equal probability \cite{syswerda1989uniform}. The second offspring is formed using the genes that are not selected for the first child. Uniform crossover cannot be used if $u=true$.
    
    \item (3) \textbf{Partially Mapped Crossover (PMX):} PMX is proposed for solution spaces with $u=true$ \cite{goldberg2014alleles}. First, a random segment is selected and the corresponding genes are swapped between the parents to create two intermediate offspring solutions. Then, a mapping is formed depending on the genes in the segment, and it is used to replace the duplicated genes on the intermediate offspring solutions until no duplication is left.
    
    \item (4) \textbf{Order Crossover (OX1):} OX1 is only applicable if $u=true$ \cite{oliver1987study}. A random segment from the first parent chromosome is selected, and the genes in this segment are copied to the first offspring. The remaining genes are inserted in the order in which they appear in the second parent. The second offspring is created by swapping the roles of the first and second parents.
    
    \item (5) \textbf{Cycle Crossover (CX):} CX is applicable to solution spaces with $u=true$ \cite{oliver1987study}. A cycle is formed starting from a random position in the first parent. Then the gene at the same position is chosen from the second parent, and the position of this value in the first parent is located. This process continues until it returns to the initial value. The genes in the cycle are copied from the first parent into the first offspring, and the remaining positions are filled from the second parent. The second offspring is created by swapping the roles of the parents.
\end{itemize}

\subsubsection{Mutation Operators} To increase variation, a mutation operator is applied with probability $p_m$ to solutions that survive elimination. Similar to crossover, our implementation allows combining different mutation operators by assigning a propensity value to each mutation operator. These values are used to randomly select one operator before each mutation operation. The mutation operators in the configuration space are explained below.


\begin{itemize}[itemsep=1mm, leftmargin=0mm, label={}]
    
    \item (1) \textbf{Point Mutation:} This operator iterates over the integers of a solution and randomly assigns a new value to each integer with probability $p_{pm}$, which is a hyperparameter \cite{spears1993overview}. Point mutation is not applicable if $u=true$.
    
    \item (2) \textbf{Swap Mutation:} Two genes are randomly selected and their positions are swapped \cite{sarmady2007investigation}. 
    
    \item (3) \textbf{Inversion Mutation:} A segment of the solution is randomly selected and the order of the genes in this segment is inverted \cite{fogel1990parallel}.
    
    \item (4) \textbf{Insertion Mutation:} A random gene and a random position in the solution are selected. The selected gene is removed from its original location and inserted into the selected position \cite{michalewicz1996genetic}.
\end{itemize}

\subsection{Simulated Annealing}

\begin{figure*}[t]
    \centering
    \includegraphics[width=0.7\linewidth]{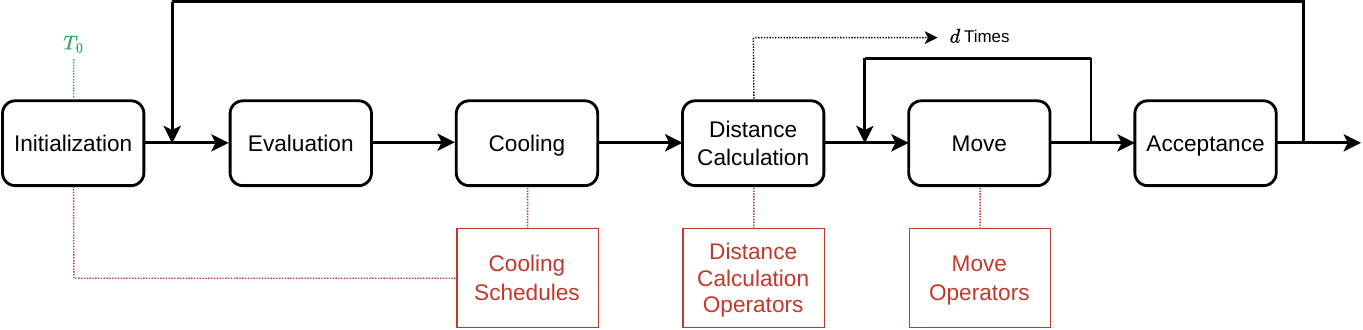}
    \caption{The structure of the simulated annealing implementation}
    \label{fig:sa}
\end{figure*}

Simulated annealing (SA) is a local search algorithm that accepts some worse solutions to escape local minima \cite{kirkpatrick1983optimization}. In the standard form of SA the acceptance probability of a neighboring solution is calculated as

\begin{equation}
        p_{acceptance}=
        \begin{cases}
            1, & \Delta C < 0 \\
            e^{-\frac{\Delta C}{T}}, & otherwise,
        \end{cases}
        \label{eq:acceptance}
    \end{equation}
where $\Delta C = C(x')-C(x)$, $x$ is the current solution, $x'$ is the neighboring solution, and $T$ is the temperature value. $T$ is gradually decreased to favor exploitation over exploration in later iterations. Figure~\ref{fig:sa} shows the structure of the SA implementation in our framework, indicating the hyperparameters (green) and operator sets (red) in the configuration space. In the following, we explain the components of the SA configuration space.

\subsubsection{General Hyperparameters} The only general hyperparameter used in SA is the initial temperature $T_0$.

\subsubsection{Cooling Schedules} 

At the beginning of each iteration, the temperature value ($T$) is updated according to a cooling schedule. The framework includes four cooling schedules.

\begin{itemize}[itemsep=1mm, leftmargin=0mm, label={}]
    \item (1) \textbf{Constant Temperature:} A constant temperature ($T_0$) is used for the entire algorithm, similarly to the Metropolis algorithm (MA) \cite{metropolis1953equation}.
    
    \item (2) \textbf{Exponential Cooling:} This schedule is also proposed in the original SA paper \cite{kirkpatrick1983optimization}. At each iteration, the temperature is multiplied by a constant value $0<c<1$.

    \item (3) \textbf{Slow Cooling:} In iteration $i$, $T$ is calculated as $T=T_0/\log(i+1)$. This schedule is used in Gaussian simulated annealing (GSA) \cite{geman1984stochastic}.

    \item (4) \textbf{Fast Cooling:} In iteration $i$, $T$ is calculated as $T=T_0/(i+1)$. This schedule is used in fast simulated annealing (FSA) \cite{szu1987fast}.

\end{itemize}

\subsubsection{Distance Calculation Operators} Our framework adapts the notion of distance calculation from continuous SA applications to discrete spaces. In this context, the distance value ($d$) is defined as the number of move operations applied during neighbor generation, and therefore takes only integer values. The operators that can be configured for distance calculation are given below.

\begin{itemize}[itemsep=1mm, leftmargin=0mm, label={}]

    \item (1) \textbf{Constant Distance:} A constant number of move operations is applied to the current solution to generate a neighbor. This is the default operator used in the original SA method \cite{kirkpatrick1983optimization}. However, in our implementation, the number of operations can also be set to a value greater than $1$.
    
    \item (2) \textbf{Uniform Distance:} In each iteration, $d$ is randomly determined using a uniform distribution, whose lower and upper bounds are set as hyperparameters. This operator is inspired by the uniform probability density used in MA \cite{metropolis1953equation}.
    
    \item (3) \textbf{Gaussian Distance:} This operator is inspired by the distance calculation in GSA \cite{geman1984stochastic}. To use it, minimum and maximum distances ($d_{min}$ and $d_{max}$) are set. The distance is calculated by adding a Gaussian random value to $d_{min}$ and rounding it to the closest integer. The standard deviation of the Gaussian distribution is set to $T$, so larger $d$ values become less likely in later iterations. If the calculated value exceeds $d_{max}$, $d$ is set to $d_{max}$.
    
    \item (4) \textbf{Cauchy Distance:} This operator is inspired by the distance calculation in FSA \cite{szu1987fast}. Similar to the Gaussian distance operator, $d_{min}$ and $d_{max}$ are set as hyperparameters. The distance is calculated by adding a Cauchy random value to $d_min$ and rounding the result to the closest integer. If this value exceeds $d_{max}$, $d$ is set to $d_{max}$. The scale parameter of the Cauchy distribution is set to $T$, so larger $d$ values become less likely in later iterations.
\end{itemize}

\subsubsection{Move Operators}

In our implementation, a neighbor is generated by applying $d$ move operations to the current solution. Multiple move operators can be combined by setting propensity values for each operator. Before each move operation, the operator to be used is randomly selected according to these propensities. The four move operators in the configuration space are listed below.

\begin{itemize}[itemsep=1mm, leftmargin=0mm, label={}]
    \item (1) \textbf{Point Move:} One of the integers in the solution string is assigned a random value. This operator is not applicable if $u=true$.
    
    \item (2) \textbf{Swap Move:} The values of two random integers in the solution string are swapped.
    
    \item (3) \textbf{Inversion Move:} A segment of the solution string is randomly selected and the order of the integers in this segment is inverted.
    
    \item (4) \textbf{Insertion Move:} One of the integers in the solution is randomly selected, removed from its location, and inserted into a random position.
\end{itemize}

\section{Experimental Evaluation}
\label{sec:experimental}

We performed a systematic analysis to identify the best algorithm families and operators for different problem types, instance characteristics, and time limits. The analysis consists of three parts. First, we evaluated algorithm families using synthetic instances. Second, we identified the operators selected by the best configurations of each family for the same set of instances. Finally, we used existing benchmark instances to investigate which findings from the first two parts of the analysis generalize beyond the controlled set of synthetic instances. In the following, we first introduce the experimental setup and then present the results of the three parts of the analysis.

\subsection{Experimental Setup}

We generated a set of 91 time limit values in the range of $[50,500]ms$. We used the analysis tool provided by STILO to determine the solution quality of the tested configurations for each time limit. These experiments were conducted on a virtual machine running Ubuntu 24.04 LTS and hosted using Proxmox VE 8.4. The physical host was a Supermicro server equipped with an AMD EPYC 9354 processor (32 cores and 64 hardware threads) and 192 GB of RAM. The virtual machine was allocated 62 vCPUs and 64 GB of RAM. To reduce the total duration of the experiments, we evaluated 60 configurations concurrently on separate threads. We measured CPU time rather than wall-clock time as a more consistent measure in this parallel computing environment.

To construct the configuration spaces used in the experiments, we considered the following hyperparameter values. All applicable combinations of these values were used in the experiments. Operator-specific hyperparameters were varied only when the corresponding operator was selected, and invalid combinations, such as those where the propensities for all crossover operators are zero for GA, were discarded.

\begin{itemize}[itemsep=1mm, leftmargin=0mm, label={}]
    \item \textbf{ACO Configurations:} For all ACO configurations, the initial pheromone, pheromone constant, and pheromone influence were all set to $1$ ($\tau_0=Q=\alpha=1$), whereas four ant counts ($an\in\{10,20,30,40\}$), nine evaporation parameters ($\rho\in\{0.1,0.2,...,0.9\}$), four heuristic influences ($\beta\in\{0,0.5,1,2\}$), and three local decay parameters ($\ell\in\{0,0.1,0.2\}$) were used. In addition, for Iteration Best, four values were considered for the number of selected tours  ($lb\in\{1,2,3,4\}$), and for Global Best, four values were considered for the number of stored tours ($gb\in\{1,2,3,4\}$).
    \item \textbf{GA Configurations:} For GA configurations, four population sizes ($ps\in\{10,20,30,40\}$), four mating-pair counts ($mn\in\{2,4,6,8\}$), three elite counts ($en\in\{0,1,2\}$), and three mutation probabilities ($p_m\in\{0.1,0.3,0.5\}$) were used. For crossover and mutation operators, we used every combination where the propensity of each operator is either 0 or 1. In addition, we used three tournament sizes ($t\in\{2,4,8\}$) for Tournament selection, three k values ($k\in\{1,2,3\}$) for K-Point crossover, and two point mutation probabilities ($p_{pm}\in\{0.01,0.02\}$) for Point mutation.
    \item \textbf{SA Configurations:} For SA configurations, we used seven initial temperatures ($T_0\in\{0.8, 1, 1.2, 1.5,2,3,4\}$). For move operators, we used every combination where the propensity of each operator is either 0 or 1. We also used three cooling parameters ($c\in\{0.999, 0.9999, 0.99999\}$) for Exponential cooling, three constant distance values ($d\in\{1,2,3\}$) for Constant distance, and three minimum distance values ($d_{min}\in\{1,2,3\}$) and five maximum distance values ($d_{max}\in\{2,4,6,8,10\}$) for the Uniform, Gaussian, and Cauchy distance calculation operators. 
\end{itemize}

For the synthetic analysis, we defined different instance classes for each problem, and generated 10 random instances for each class, each of which was solved once by each configuration. For each configuration and time limit, we defined the solution quality as the average objective value obtained over 10 instances in the corresponding class. The parameter values that we used to generate the synthetic instances are listed below. Here $N(\mu,\sigma)$ denotes a normal distribution with mean $\mu$ and standard deviation $\sigma$. Job lengths, weights, and distances are expected to be positive continuous values. This is ensured by sampling until a positive value is obtained.

\begin{itemize}[itemsep=1mm, leftmargin=0mm, label={}]
    \item \textbf{IMSP Parameters:} For IMSP, we combined two numbers of jobs ($n \in \{50,100\}$), two numbers of machines ($m \in \{4,8\}$), and two normal distributions for job lengths ($N(100,10)$ and $N(100,25)$), resulting in eight instance classes. 
    \item \textbf{MCP Parameters:} For MCP, we combined two numbers of vertices ($n \in \{50,100\}$) and three edge densities ($d \in \{0.25,0.50,0.75\}$), resulting in six instance classes. During random instance generation, $d$ specifies the probability that an edge is included between each pair of vertices.
    \item \textbf{WMCP Parameters:} For WMCP, we combined two numbers of vertices ($n \in \{50,100\}$) and two normal distributions for edge weights ($N(100,10)$ and $N(100,25)$), yielding four instance classes. In WMCP instance classes, we used fully connected graphs unlike MCP, and edge weights were determined randomly.
    \item \textbf{TSP Parameters:} For TSP, we combined two numbers of cities ($n \in \{50,100\}$) and two normal distributions for distances between cities ($N(100,10)$ and $N(100,25)$), yielding four instance classes. We only considered symmetric TSP and assigned the same distance to both directions between each pair of cities.
    \item \textbf{TxnSP Parameters:} We combined two numbers of jobs ($n \in \{50,100\}$), two numbers of machines ($m \in \{4,8\}$), three conflict probabilities ($cp \in \{0.25,0.50,0.75\}$), and two normal distributions for job lengths ($N(100,10)$ and $N(100,25)$), yielding 24 instance classes. During random instance generation, $cp$ specifies the conflict probability of each pair of jobs.
\end{itemize}

For IMSP, we used four instances with $n\in \{120,217\}$ and $m\in\{4,7\}$ from the set of instances introduced by Schreiber and Korf~\cite{schreiber2014cached}. For MCP, we used the sg3dl051000, sg3dl052000, toursg3-8, and tourspm3-8-50 instances from the Max-Cut benchmark collection of Martí et al.~\cite{marti2009advanced}. All four instances contain edge weights other than 0 and 1, and therefore can be classified as WMCP. The first two contain 125 vertices and the edge weights are restricted to $\{-1,0,1\}$. The latter two contain 512 vertices and have a wider range of the edge weights. For TSP, we used the Berlin52, Eil76, Rd100, and Tsp225 instances from
TSPLIB~\cite{reinelt1991tsplib}, containing 52, 76, 100, and 225 cities, respectively. Since no established benchmark instances are available for TxnSP, we generated instances using the TPC-C workload, which is a widely used database benchmark for OLTP systems \cite{tpc2010tpcc}. We generated transactions from this workload and estimated their processing times by executing them using SQLite and measuring the execution times. We constructed the conflict matrices assuming that two transactions conflict if both perform a write operation to the same database table. We generated four instances with $n\in \{100,200\}$ and $m\in\{4,8\}$. We solved each benchmark instance 10 times using each solver configuration, and used the average of the objective values over these runs as the measure of solution quality.

\subsection{Evaluation of the Algorithm Families}
\label{ssec:algorithmevaluation}

\begin{figure*}
    \centering
    \begin{subfigure}{0.32\textwidth}
        \includegraphics[width=\textwidth]{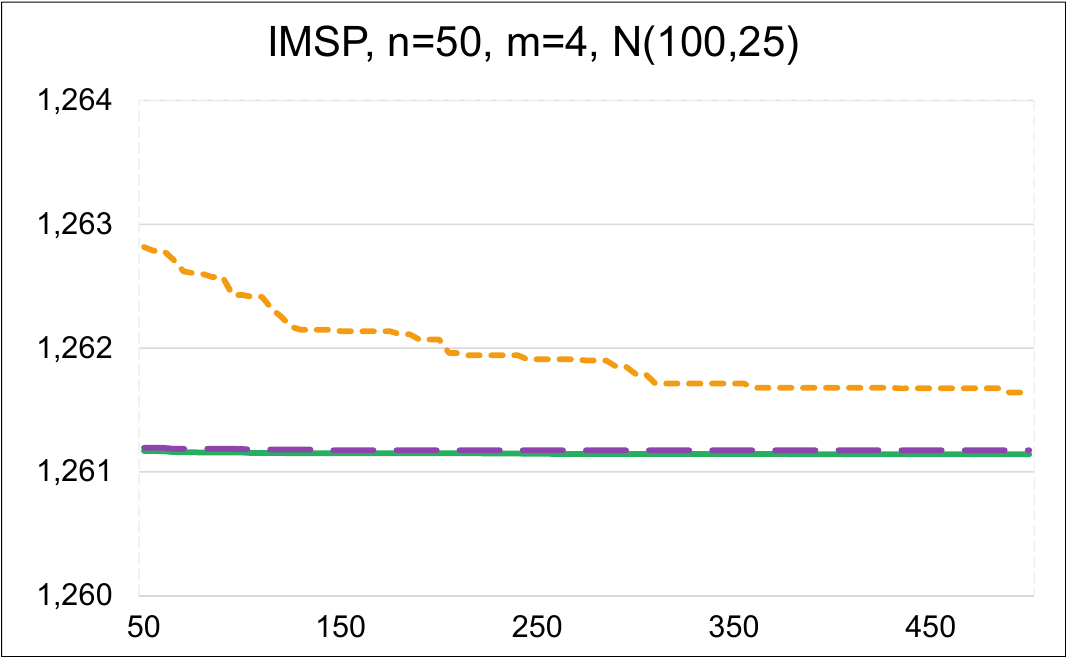}
    \end{subfigure}
    \hfill
    \begin{subfigure}{0.32\textwidth}
        \includegraphics[width=\textwidth]{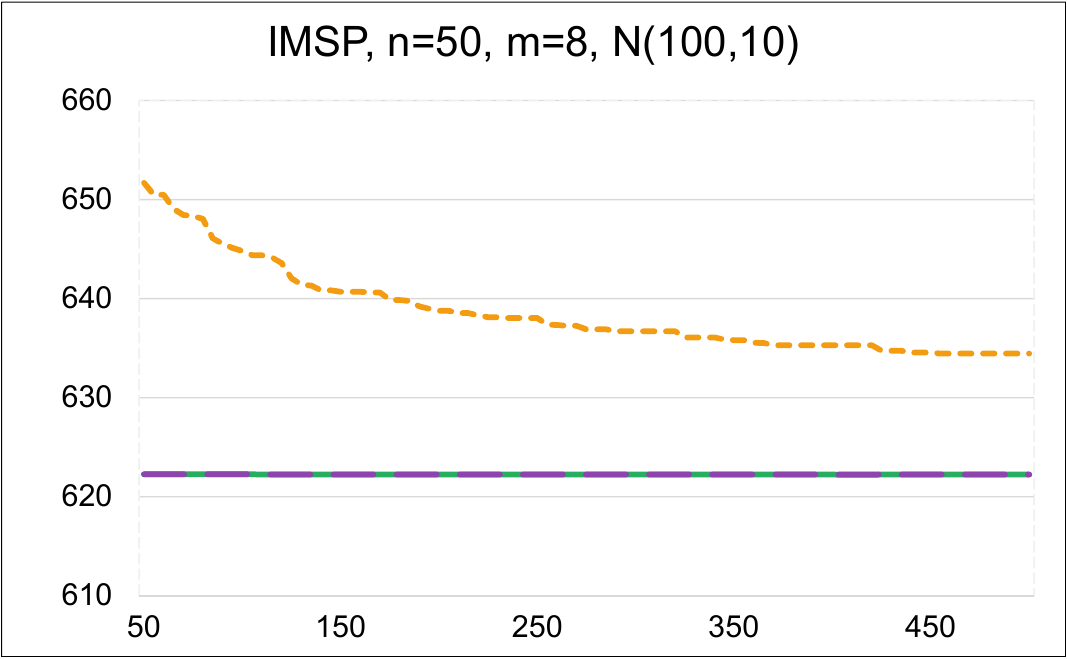}
    \end{subfigure}
    \hfill
    \begin{subfigure}{0.32\textwidth}
        \includegraphics[width=\textwidth]{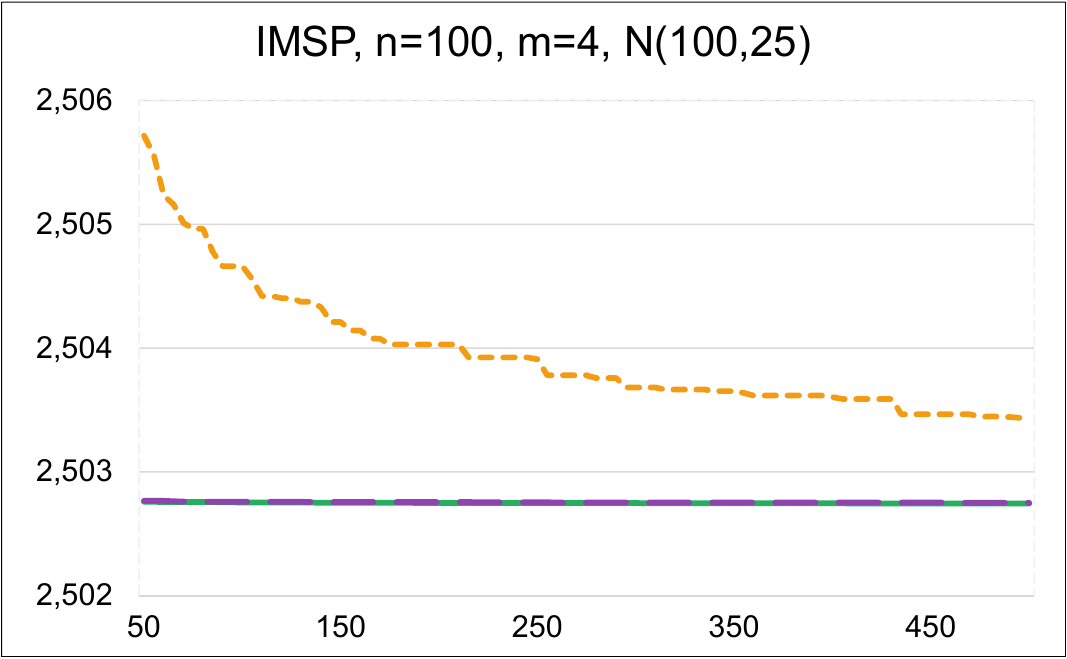}
    \end{subfigure}
    \begin{subfigure}{0.32\textwidth}
        \includegraphics[width=\textwidth]{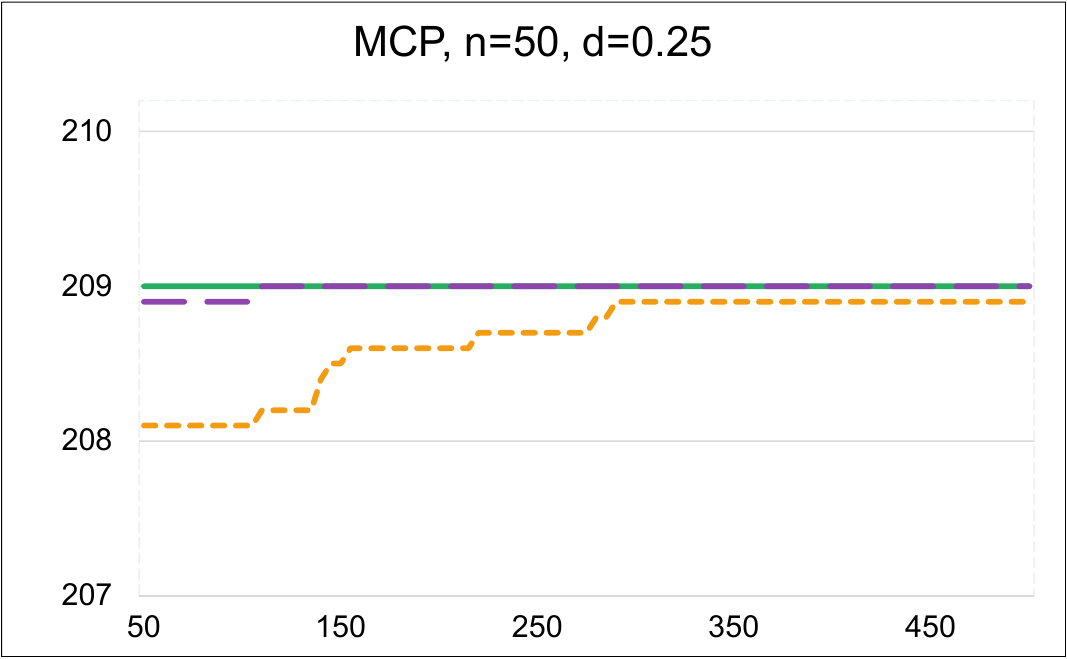}
    \end{subfigure}
    \hfill
    \begin{subfigure}{0.32\textwidth}
        \includegraphics[width=\textwidth]{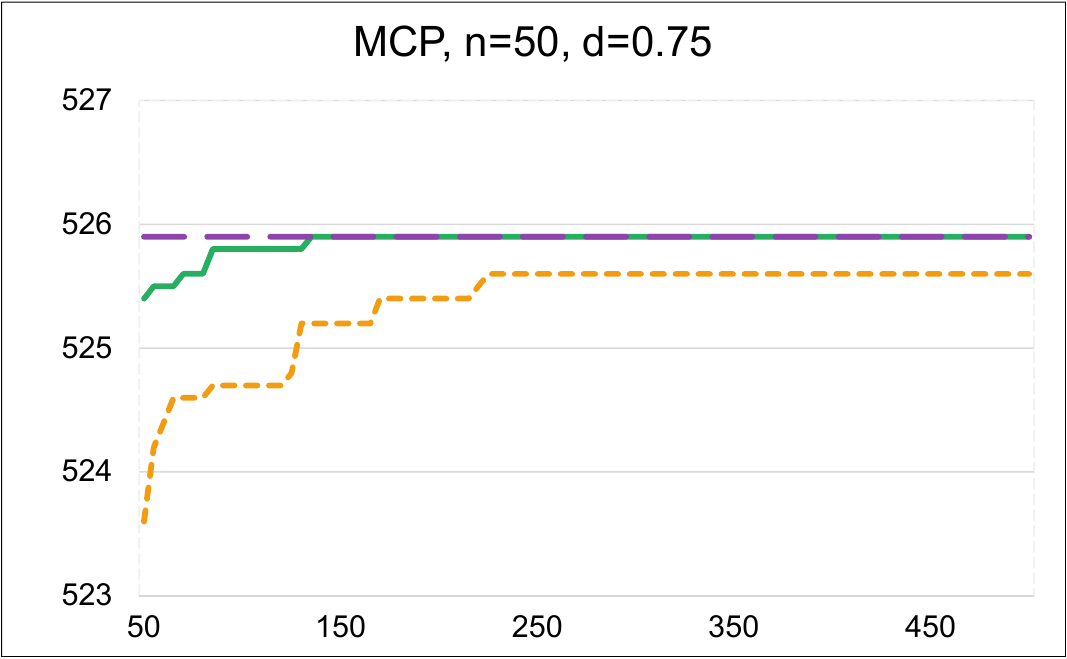}
    \end{subfigure}
    \hfill
    \begin{subfigure}{0.32\textwidth}
        \includegraphics[width=\textwidth]{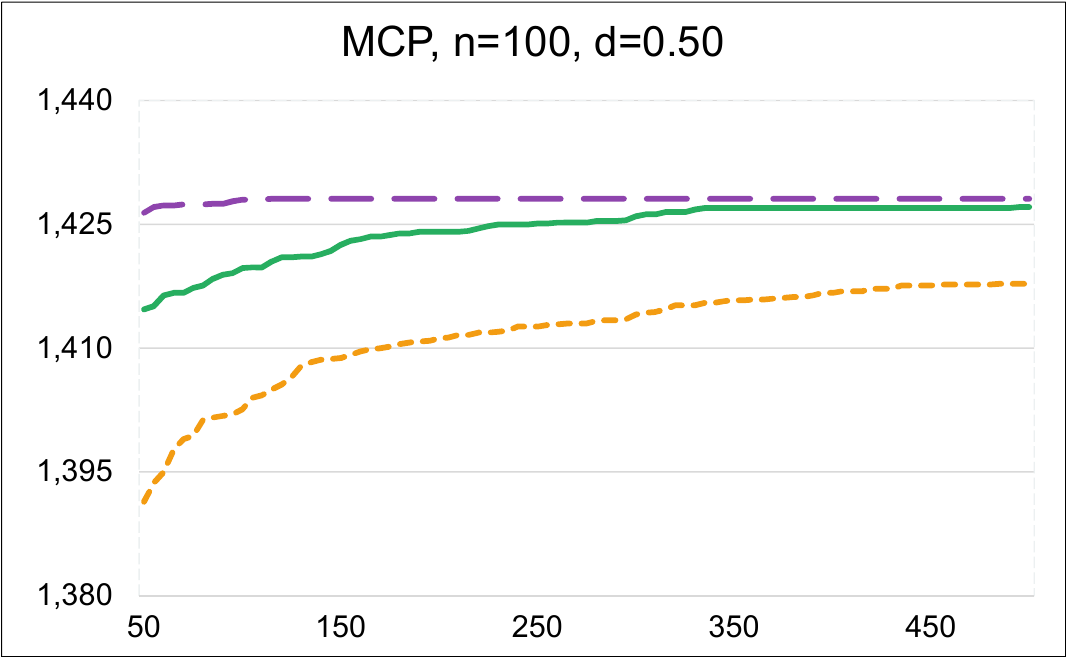}
    \end{subfigure}
    \begin{subfigure}{0.32\textwidth}
        \includegraphics[width=\textwidth]{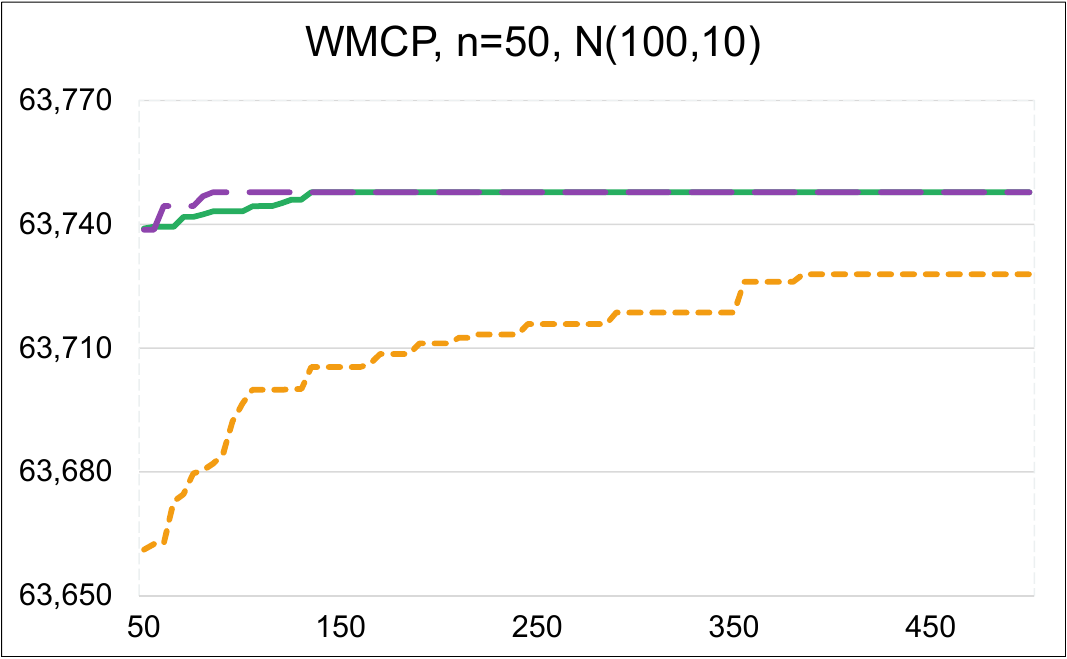}
    \end{subfigure}
    \hfill
    \begin{subfigure}{0.32\textwidth}
        \includegraphics[width=\textwidth]{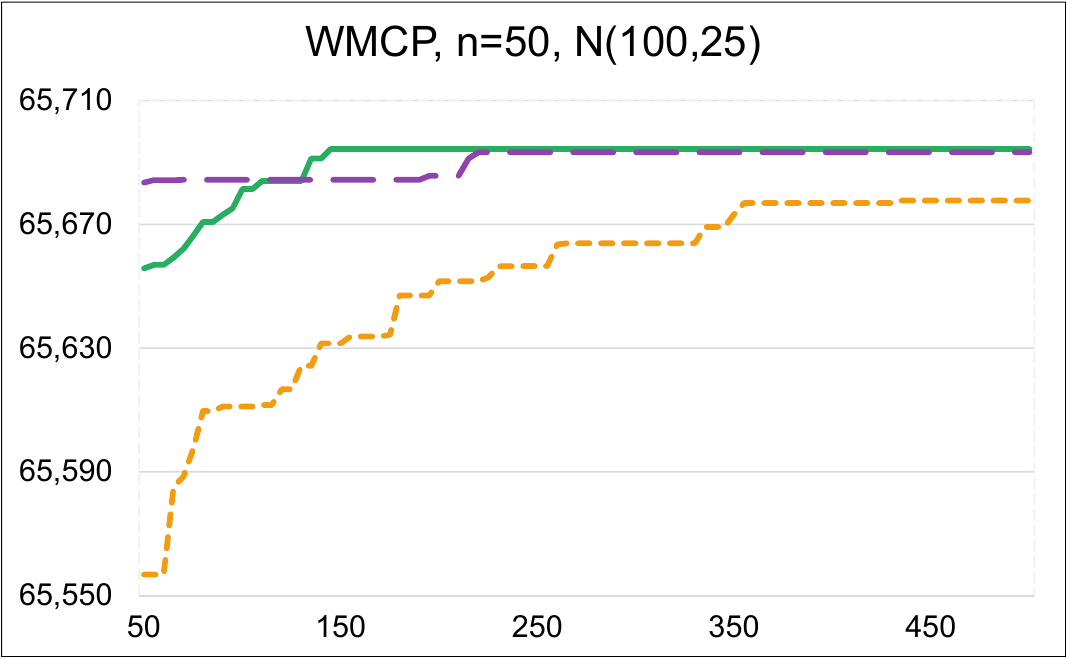}
    \end{subfigure}
    \hfill
    \begin{subfigure}{0.32\textwidth}
        \includegraphics[width=\textwidth]{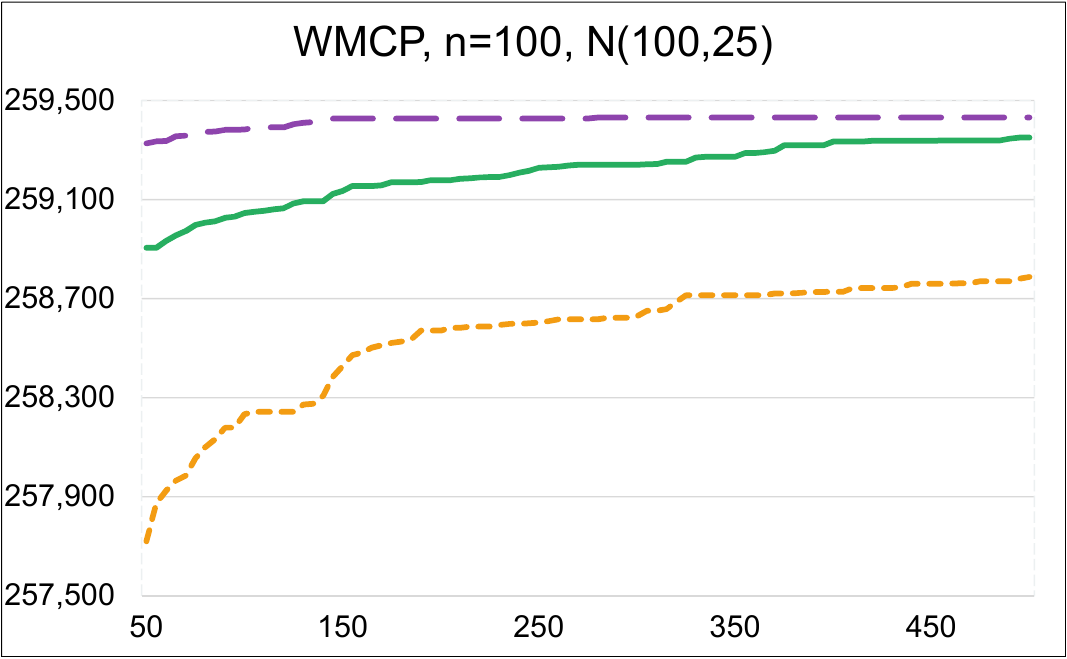}
    \end{subfigure}
    \begin{subfigure}{0.32\textwidth}
        \includegraphics[width=\textwidth]{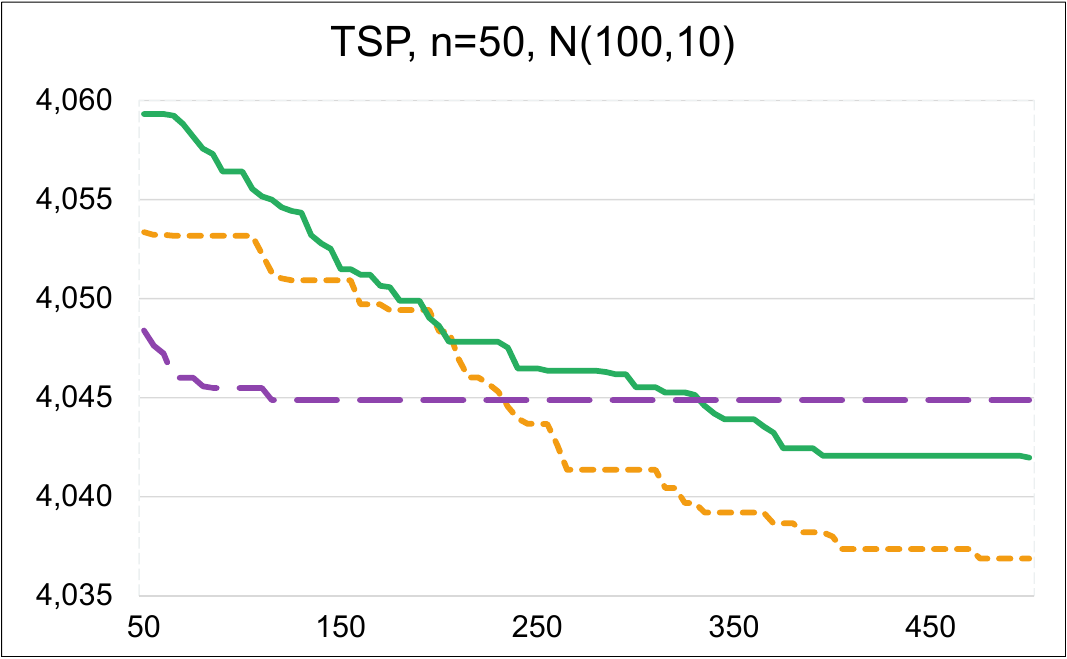}
    \end{subfigure}
    \hfill
    \begin{subfigure}{0.32\textwidth}
        \includegraphics[width=\textwidth]{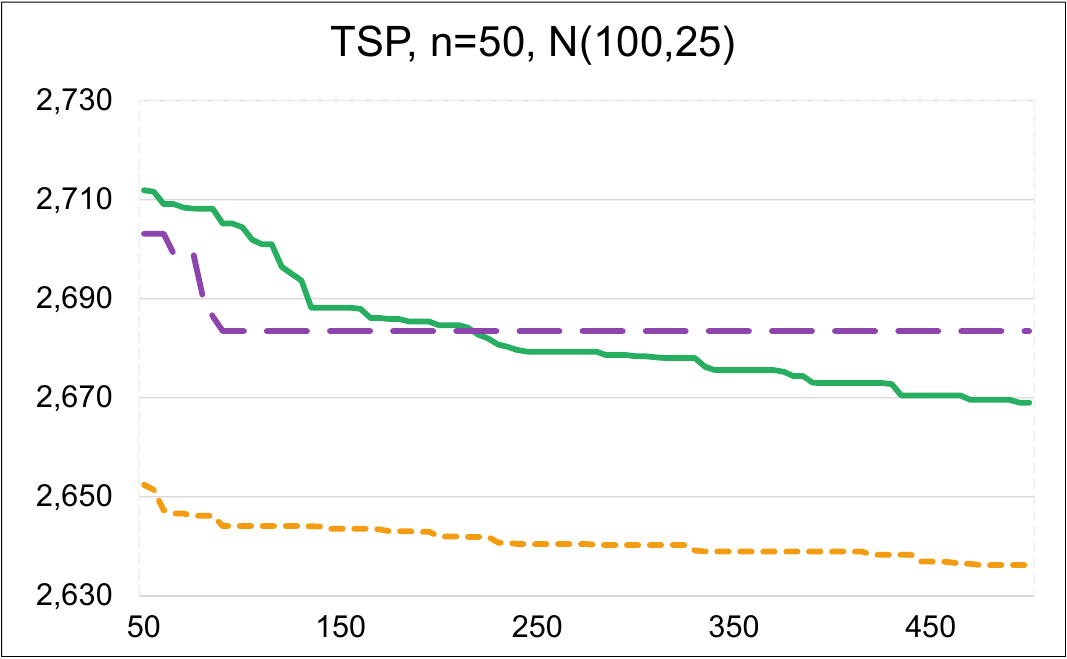}
    \end{subfigure}
    \hfill
    \begin{subfigure}{0.32\textwidth}
        \includegraphics[width=\textwidth]{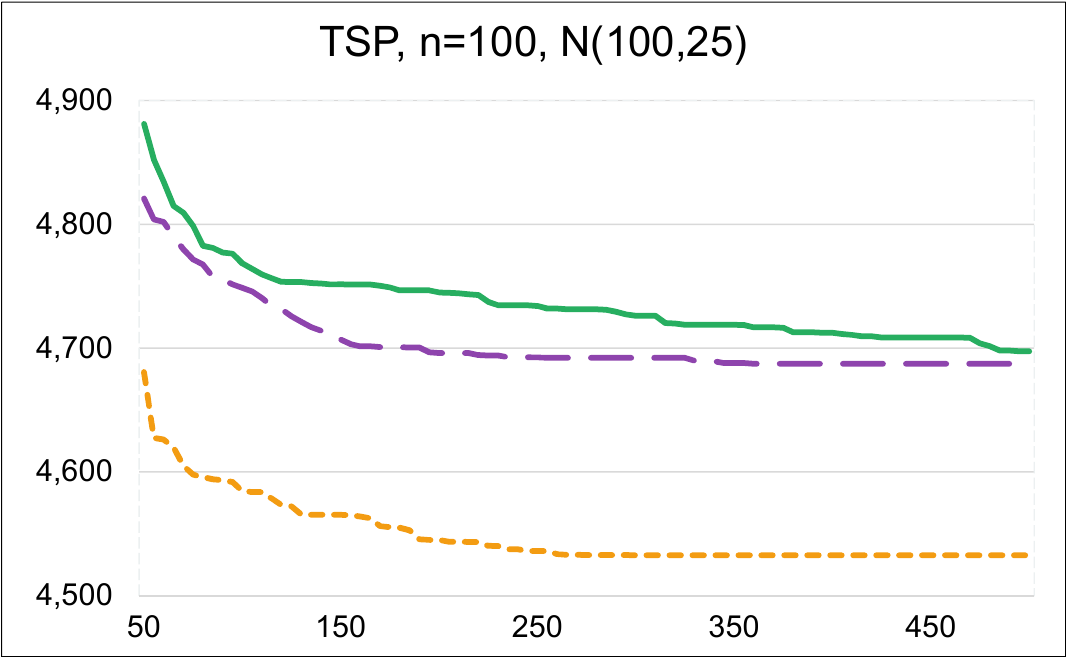}
    \end{subfigure}
    \begin{subfigure}{0.32\textwidth}
        \includegraphics[width=\textwidth]{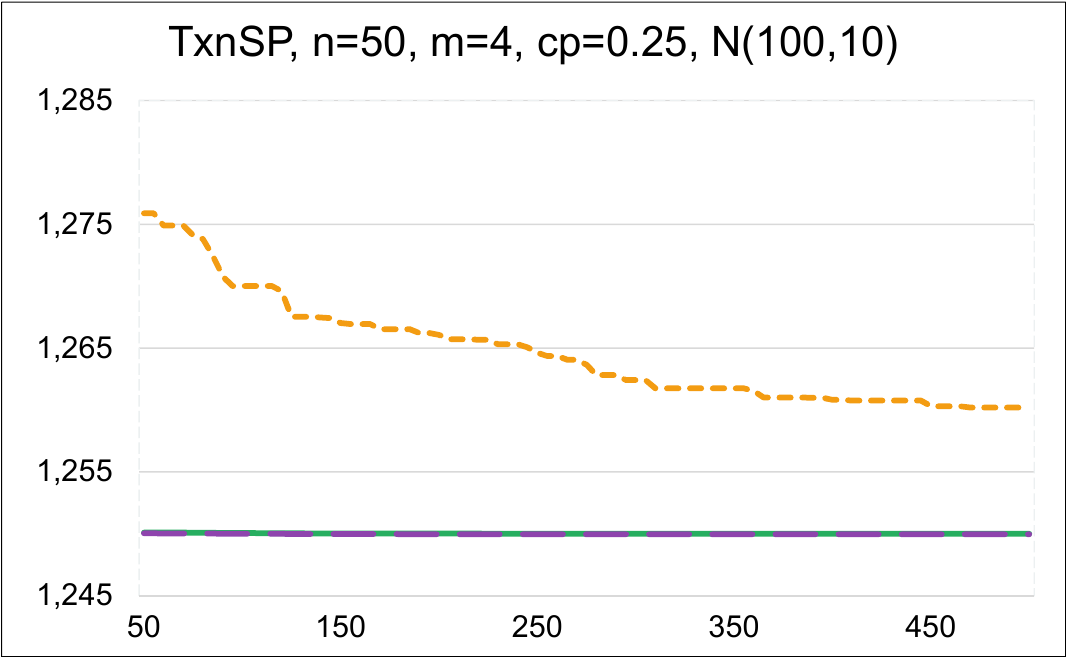}
    \end{subfigure}
    \hfill
    \begin{subfigure}{0.32\textwidth}
        \includegraphics[width=\textwidth]{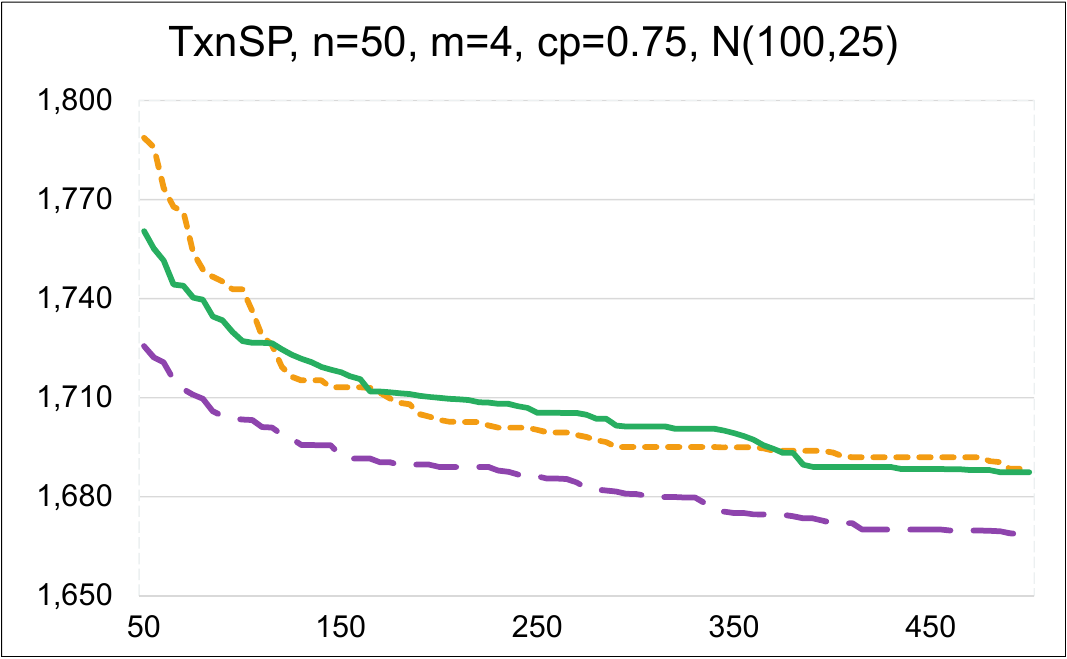}
    \end{subfigure}
    \hfill
    \begin{subfigure}{0.32\textwidth}
        \includegraphics[width=\textwidth]{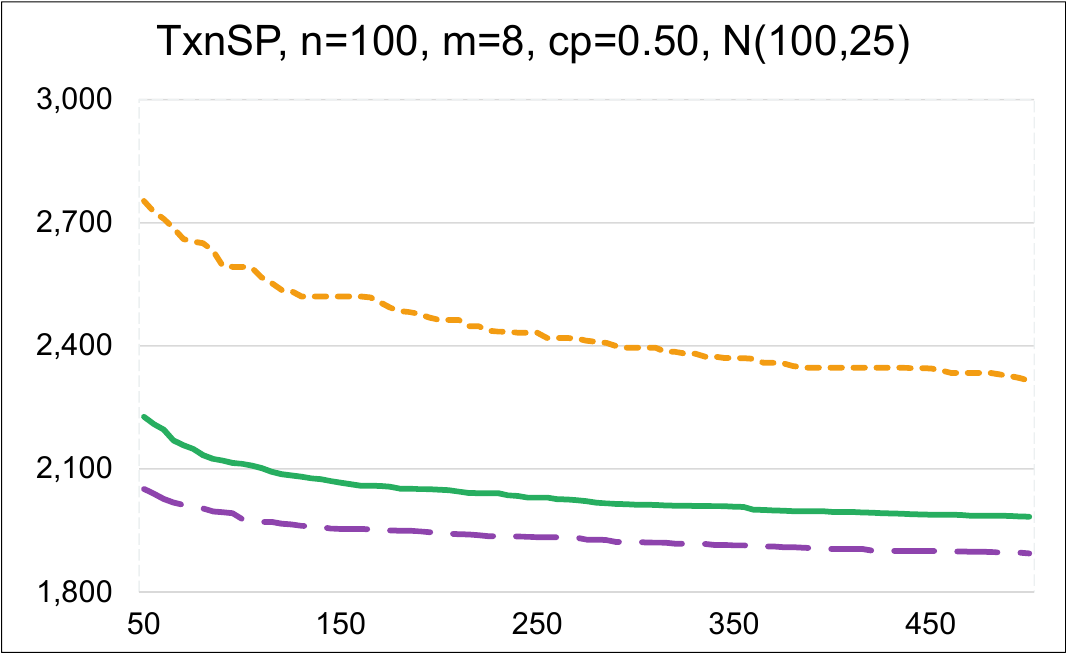}
    \end{subfigure}
    
    \caption{Comparison of the algorithm families for various time limits. The horizontal axes show the time limits in milliseconds and the vertical axes show the average objective values achieved by the best-performing configuration of each family under the associated time limit. For IMSP, TSP, and TxnSP, a lower objective value is better, whereas for MCP and WMCP, a higher objective value is better. The best-performing configurations of {\color{flatorange} ACO}, {\color{flatgreen} GA}, and {\color{flatpurple} SA} are represented by {\color{flatorange} orange short-dashed}, {\color{flatgreen} green solid} and {\color{flatpurple} purple long-dashed} lines, respectively.}
    \label{fig:algorithmfamilies}
\end{figure*}

To evaluate the algorithm families, we determined the best-performing configuration of each family separately for each instance class and time limit, based on the average solution quality over the 10 instances in that class. Figure~\ref{fig:algorithmfamilies} compares the average solution qualities achieved by these configurations under varying time limits. Due to space limitations, we cannot include the graphs for every instance class of every problem. Instead, we present representative graphs that illustrate the general patterns observed throughout the complete set of results.

For all IMSP instance classes, the best GA and SA configurations achieved similar solution quality, with GA performing slightly better, especially for small instances ($n=50$). Both families outperformed ACO for every tested instance class and time limit. ACO performed worse than GA and SA for all tested instance classes of MCP, similarly to IMSP. For small instances ($n=50$), SA and GA achieved similar solution qualities for most time limits, while their relative performance under short time limits depended on the edge density. For low densities ($d\in \{0.25,0.50\}$), GA found better solutions under short time limits, whereas SA performed better for high density ($d=0.75$) under short time limits. For large instances ($n=100$), SA consistently outperformed GA over the entire range of time limits.
Similar behavior was observed for WMCP. For small instances, GA and SA achieved similar solution qualities for most time limits, whereas SA outperformed GA for large instances. ACO was not competitive with the other algorithm families for any tested instance class.
In contrast, for TSP, ACO was the best-performing algorithm family. For instance classes with the larger standard deviation ($N(100,25)$), ACO achieved significantly better solution quality than GA and SA over the entire range of time limits. For instance classes with smaller standard deviation, GA and SA were able to occasionally achieve the best solution qualities under short to moderate time limits, while ACO became the best-performing algorithm family as the time limit increased.
For TxnSP instance classes, SA achieved the best solution qualities. GA remained competitive for some instance classes, particularly with parameters $m=4$ and $cp=0.25$. ACO was not able to provide a comparable solution quality for any tested instance class of TxnSP.

From the analysis results, we observed certain patterns in the behavior of the algorithm families under strict time limits. The first conclusion we draw is that the most effective algorithm family for a given setting depends on the instance characteristics and the allowed execution time, in addition to the general problem class. We observed that even when an algorithm family outperforms the others for most cases, some other family could be a better choice for certain instance classes and time limits. The results show that SA is effective for at least some instance classes of all tested problems. GA was mostly competitive with SA and occasionally exceeded its performance for small instances or short time limits, showing that it is a viable choice for certain optimization settings. Within the tested problem set, ACO was less effective than GA and SA as a general-purpose solver, but was highly effective for TSP.

\subsection{Evaluation of the Operators}

To evaluate the operators, we analyzed the operators utilized by the best-performing configurations of each algorithm family for each instance class and time limit, as determined in Section~\ref{ssec:algorithmevaluation}. The results revealed the operator preferences for different problems and the changes in the effectiveness of operators depending on instance parameters and computational budget. 

As observed in Section~\ref{ssec:algorithmevaluation}, for some instance classes, the best observed average objective value found by an algorithm family stabilized after a certain time limit and did not improve further in the tested range. After stabilization, changes in the selected best-performing configuration become less informative, since multiple configurations are more likely to achieve similar average objective values once no further improvement is observed. To account for this factor, we determined a \textit{stabilization time limit} for each algorithm-family and instance-class pair, defined as the largest tested time limit at which an improvement in the average objective value was observed. Behavior before the stabilization time limit is given greater emphasis in the analysis, whereas changes in operator selection after this limit are interpreted with greater caution.

In the following, we present the operator analysis for each algorithm family.

\subsubsection{ACO Operators}

\begin{figure*}
    \begin{subfigure}{0.32\textwidth}
        \includegraphics[width=\textwidth]{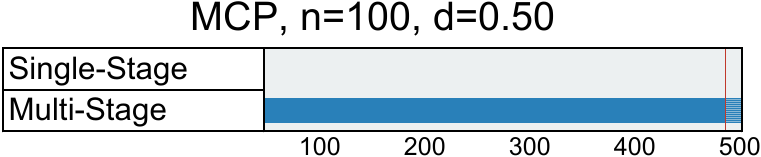}
    \end{subfigure}
    \hfill
    \begin{subfigure}{0.32\textwidth}
        \includegraphics[width=\textwidth]{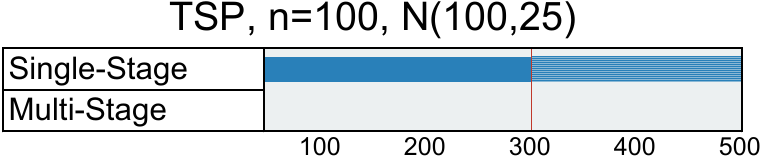}
    \end{subfigure}
    \hfill
    \begin{subfigure}{0.32\textwidth}
        \includegraphics[width=\textwidth]{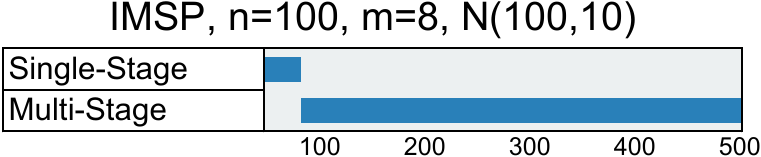}
    \end{subfigure}
    \caption{Utilization of the graph structures by the best-performing ACO configurations. The horizontal axes show the time limits in milliseconds and the blue segments mark the time limits for which each structure is selected. The vertical red line indicates the stabilization time limit.}
    \label{fig:acograph}
\end{figure*}

\begin{itemize}[itemsep=1mm, leftmargin=0mm, label={}]
    \item \textbf{Graph Structures:} The results show that the effectiveness of the graph structures strongly depends on the structure of the search space. For the problems with $u=true$ (TSP and TxnSP), Single-Stage graph structure was selected throughout the tested time limit range. These problems have large $sn$ values, making the computational overhead of Multi-Stage graphs very significant for optimization under strict time limits. In contrast, Multi-Stage structure was selected for almost the entire time limit range of the problems with $u=false$. Since these problems have small $sn$ values, a Single-Stage graph structure can store only a very limited amount of information. The only exception occurred for a single instance class of IMSP, where Single-Stage graph was used for the shortest time limits before stabilization. This result shows that when the computational budget is limited, graph structures with lower overhead could be preferred despite their lower representation capacity. Figure~\ref{fig:acograph} illustrates these patterns.

\begin{figure*}
    \begin{subfigure}{0.32\textwidth}
        \includegraphics[width=\textwidth]{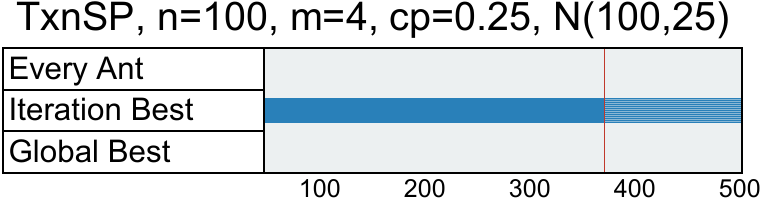}
    \end{subfigure}
    \hfill
    \begin{subfigure}{0.32\textwidth}
        \includegraphics[width=\textwidth]{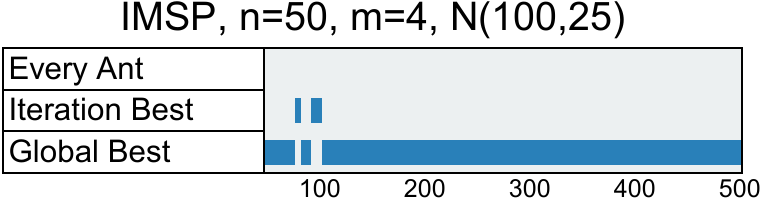}
    \end{subfigure}
    \hfill
    \begin{subfigure}{0.32\textwidth}
        \includegraphics[width=\textwidth]{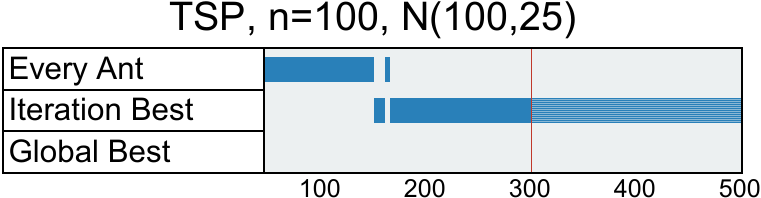}
    \end{subfigure}
    \caption{Utilization of the tour selection operators by the best-performing ACO configurations. The horizontal axes show the time limits in milliseconds and the blue segments mark the time limits for which each operator is selected. The vertical red line indicates the stabilization time limit.}
    \label{fig:acotour}
\end{figure*}

    \item \textbf{Tour Selection Operators:} Iteration Best was the most favorable tour selection operator, being used over the majority of the time limit range for most instance classes. Global Best was also selected for portions of the time limit range in several instance classes, and was particularly prominent for IMSP instances with high standard deviation. Every Ant tour selection was only used under short time limits for a single TSP instance class, and was not used elsewhere. These results are shown in Figure~\ref{fig:acotour}.

\begin{figure*}
    \begin{subfigure}{0.32\textwidth}
        \includegraphics[width=\textwidth]{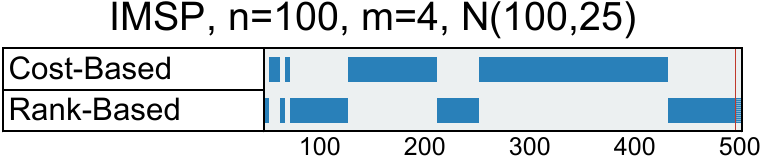}
    \end{subfigure}
    \hfill
    \begin{subfigure}{0.32\textwidth}
        \includegraphics[width=\textwidth]{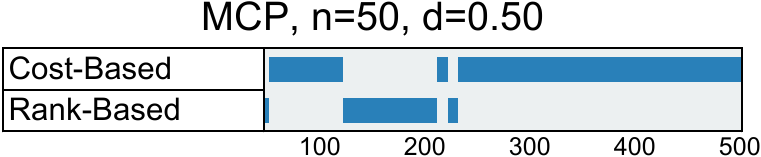}
    \end{subfigure}
    \hfill
    \begin{subfigure}{0.32\textwidth}
        \includegraphics[width=\textwidth]{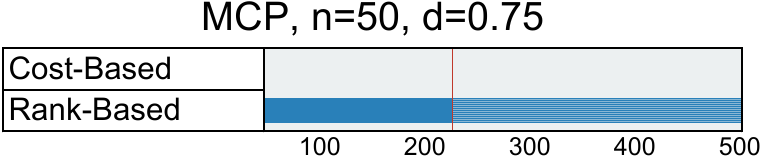}
    \end{subfigure}
    
    \vspace{2mm}
    
    \begin{subfigure}{0.32\textwidth}
        \includegraphics[width=\textwidth]{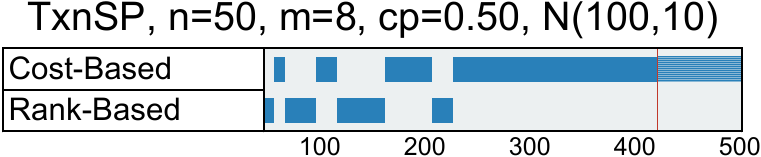}
    \end{subfigure}
    \hfill
    \begin{subfigure}{0.32\textwidth}
        \includegraphics[width=\textwidth]{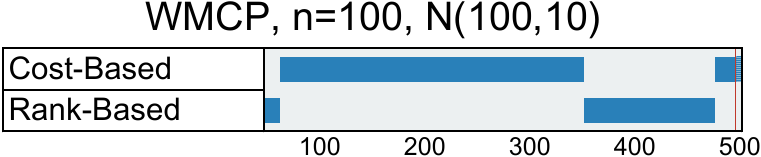}
    \end{subfigure}
    \hfill
    \begin{subfigure}{0.32\textwidth}
        \includegraphics[width=\textwidth]{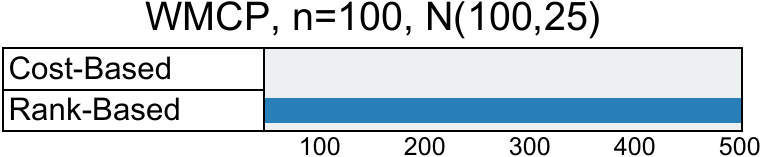}
    \end{subfigure}
    \caption{Utilization of the pheromone calculation operators by the best-performing ACO configurations. The horizontal axes show the time limits in milliseconds and the blue segments mark the time limits for which each operator is selected. The vertical red line indicates the stabilization time limit.}
    \label{fig:acocalculation}
\end{figure*}

    \item \textbf{Pheromone Calculation Operators:} In our experiments, pheromone calculation was the most sensitive ACO operator choice to instance characteristics and time limits. For most instance classes, the Cost-Based and Rank-Based calculation operators alternated over the range of time limits, without either being the dominant choice. MCP instance classes with high edge densities ($d=0.75$) and WMCP instance classes with high standard deviation ($N(100,25)$) were the exceptions, as Rank-Based calculation was selected exclusively throughout the tested time limit range. TSP and TxnSP did not show a consistent preference, as some instance classes selected Cost-Based calculation, others Rank-Based calculation, while others switched between the two operators for different time limits. Figure~\ref{fig:acocalculation} presents some representative examples.

\begin{figure*}
    \begin{subfigure}{0.32\textwidth}
        \includegraphics[width=\textwidth]{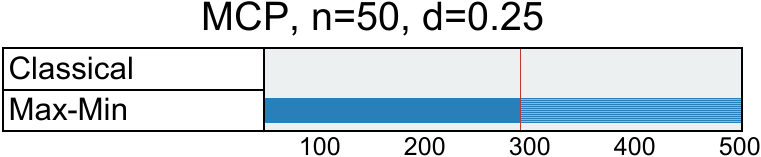}
    \end{subfigure}
    \hfill
    \begin{subfigure}{0.32\textwidth}
        \includegraphics[width=\textwidth]{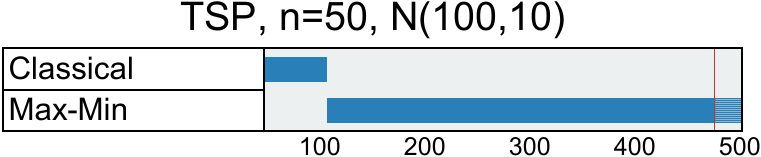}
    \end{subfigure}
    \hfill
    \begin{subfigure}{0.32\textwidth}
        \includegraphics[width=\textwidth]{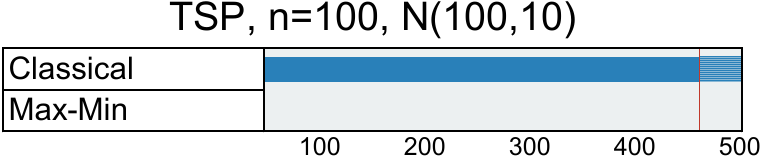}
    \end{subfigure}

    \vspace{2mm}
    
    \begin{subfigure}{0.32\textwidth}
        \includegraphics[width=\textwidth]{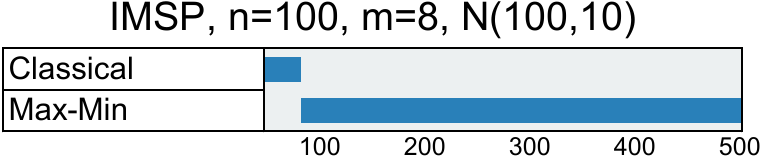}
    \end{subfigure}
    \hfill
    \begin{subfigure}{0.32\textwidth}
        \includegraphics[width=\textwidth]{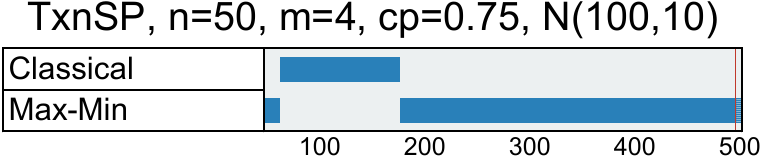}
    \end{subfigure}
    \hfill
    \begin{subfigure}{0.32\textwidth}
        \includegraphics[width=\textwidth]{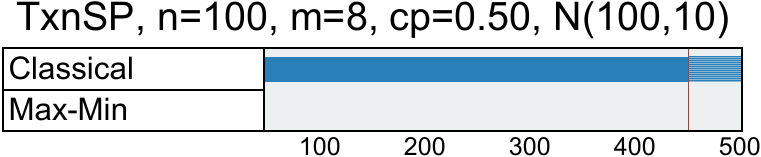}
    \end{subfigure}
    \caption{Utilization of the pheromone update operators by the best-performing ACO configurations. The horizontal axes show the time limits in milliseconds and the blue segments mark the time limits for which each operator is selected. The vertical red line indicates the stabilization time limit.}
    \label{fig:acoupdate}
\end{figure*}

    \begin{figure*}
    \begin{subfigure}{0.32\textwidth}
        \includegraphics[width=\textwidth]{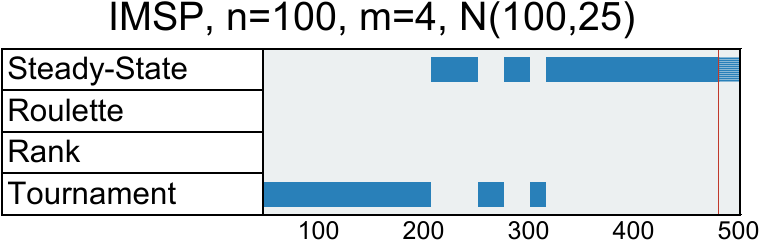}
    \end{subfigure}
    \hfill
    \begin{subfigure}{0.32\textwidth}
        \includegraphics[width=\textwidth]{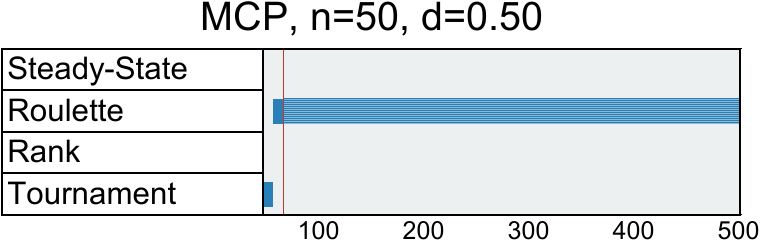}
    \end{subfigure}
    \hfill
    \begin{subfigure}{0.32\textwidth}
        \includegraphics[width=\textwidth]{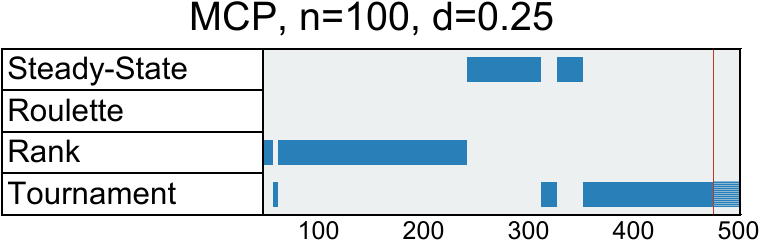}
    \end{subfigure}

    \vspace{2mm}
    
    \begin{subfigure}{0.32\textwidth}
        \includegraphics[width=\textwidth]{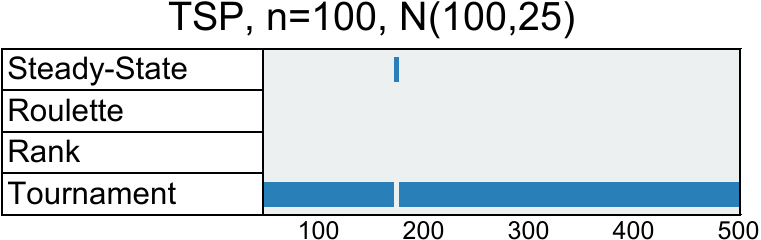}
    \end{subfigure}
    \hfill
    \begin{subfigure}{0.32\textwidth}
        \includegraphics[width=\textwidth]{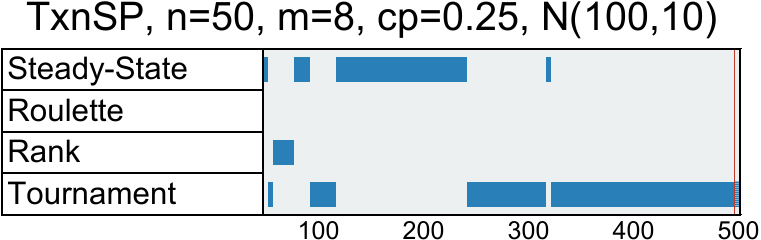}
    \end{subfigure}
    \hfill
    \begin{subfigure}{0.32\textwidth}
        \includegraphics[width=\textwidth]{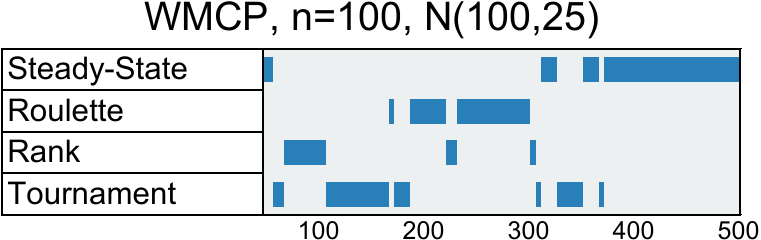}
    \end{subfigure}
    \caption{Utilization of the selection operators by the best-performing GA configurations. The horizontal axes show the time limits in milliseconds and the blue segments mark the time limits for which each operator is selected. The vertical red line indicates the stabilization time limit.}
    \label{fig:gaselection}
\end{figure*}

    \item \textbf{Pheromone Update Operators:} Pheromone update operators showed strong problem dependence. The Max-Min update operator was used for almost the entire time limit range of all IMSP, MCP, and WMCP classes, with Classical update being used only briefly for a single IMSP instance class. In contrast, Classical update dominated most TxnSP instance classes, although Max-Min update was competitive for some $n=50$ classes, particularly those with $cp=0.75$. For TSP, the preference depended on the instance size. For small TSP instances, Max-Min update was selected more often, whereas for large instances, Classical update was used more frequently. Figure~\ref{fig:acoupdate} displays these results.

\end{itemize}

\subsubsection{GA Operators}

\begin{itemize}[itemsep=1mm, leftmargin=0mm, label={}]

    \item \textbf{Selection Operators:} The results show that the effectiveness of different selection operators is highly problem-dependent. For IMSP instances, the best configurations almost exclusively used Steady-State and Tournament selection, alternating between them for different time limits. Although small MCP instances stabilized very early, Roulette was used extensively, with Tournament selection being used only briefly for a single instance class. Large MCP instances showed greater variation, with all operators being selected for different time limits. For TSP, Tournament selection was the most frequently selected operator as it dominated all instance classes except one. Similarly, for TxnSP, Tournament selection was the most frequently used operator, followed by Steady-State selection. Rank selection was also used occasionally, but Roulette selection was never selected. There was no single dominant operator for WMCP, with the selected operator varying among time limits and instance classes. Figure~\ref{fig:gaselection} presents representative results.

\begin{figure*}
    \begin{subfigure}{0.32\textwidth}
        \includegraphics[width=\textwidth]{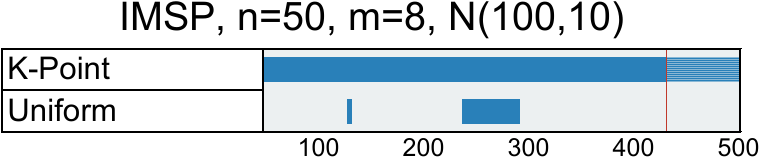}
    \end{subfigure}
    \hfill
    \begin{subfigure}{0.32\textwidth}
        \includegraphics[width=\textwidth]{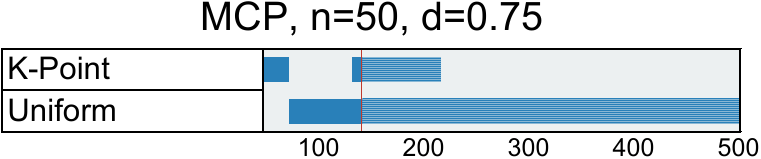}
    \end{subfigure}
    \hfill
    \begin{subfigure}{0.32\textwidth}
        \includegraphics[width=\textwidth]{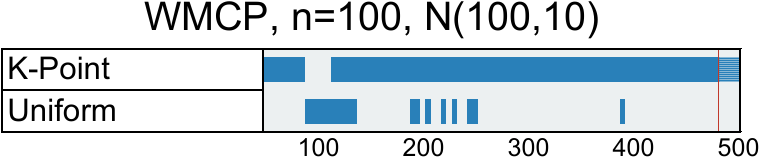}
    \end{subfigure}

    \vspace{2mm}
    
    \begin{subfigure}{0.32\textwidth}
        \includegraphics[width=\textwidth]{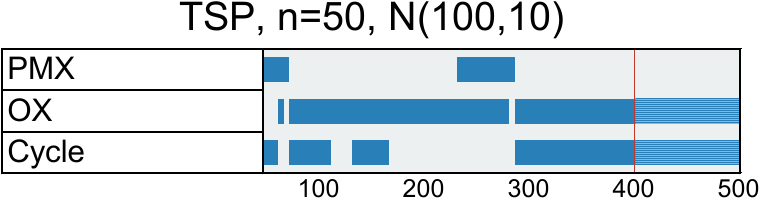}
    \end{subfigure}
    \hfill
    \begin{subfigure}{0.32\textwidth}
        \includegraphics[width=\textwidth]{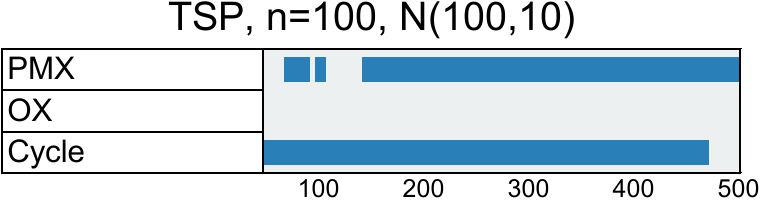}
    \end{subfigure}
    \hfill
    \begin{subfigure}{0.32\textwidth}
        \includegraphics[width=\textwidth]{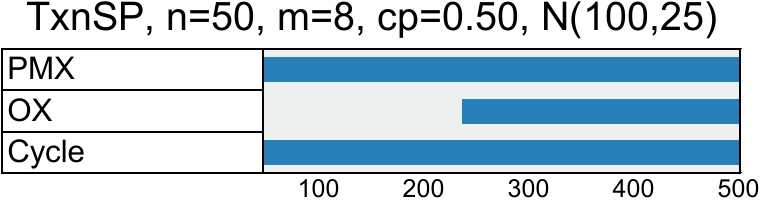}
    \end{subfigure}
    \caption{Utilization of the crossover operators by the best-performing GA configurations. The horizontal axes show the time limits in milliseconds and the blue segments mark the time limits for which each operator is selected. The vertical red line indicates the stabilization time limit.}
    \label{fig:gacrossover}
\end{figure*}

    \item \textbf{Crossover Operators:} The results for the crossover operators revealed that, for most settings, a combination of operators could be more effective than a single operator. For IMSP, K-Point crossover was used for the entire time limit range of every instance class, while Uniform crossover was occasionally added to the combination. For small MCP instances, there was no dominant operator before their early stabilization, and for large instances the relative utilization levels of K-Point and Uniform crossovers varied between instance classes. WMCP classes generally favored K-Point, with Uniform crossover being utilized for substantial portions of the time limit range, particularly for small instances before stabilization. For TSP, Cycle crossover was used extensively for all instance classes, while the utilization of other operators depended on the instance size. For $n=50$, OX crossover was selected more frequently, whereas for $n=100$, PMX crossover was more prominent. For TxnSP, each of the three crossover operators was used in at least some portion of the time limits. PMX and Cycle were selected more frequently than OX. Figure~\ref{fig:gacrossover} shows these patterns.

\begin{figure*}
    \begin{subfigure}{0.32\textwidth}
        \includegraphics[width=\textwidth]{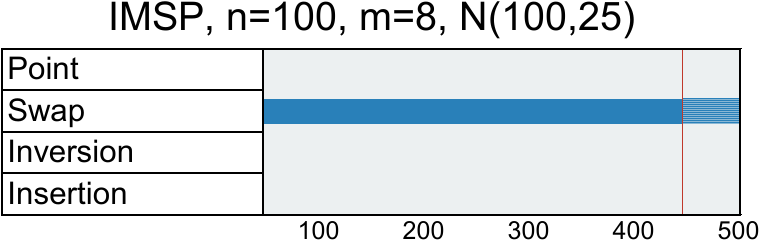}
    \end{subfigure}
    \hfill
    \begin{subfigure}{0.32\textwidth}
        \includegraphics[width=\textwidth]{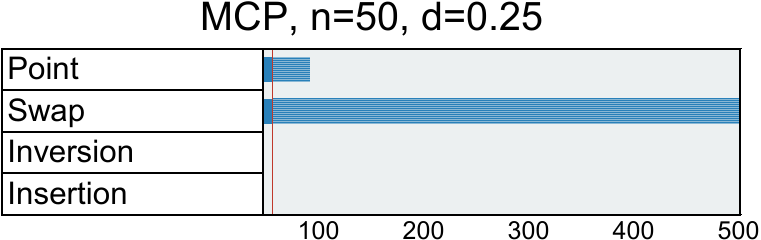}
    \end{subfigure}
    \hfill
    \begin{subfigure}{0.32\textwidth}
        \includegraphics[width=\textwidth]{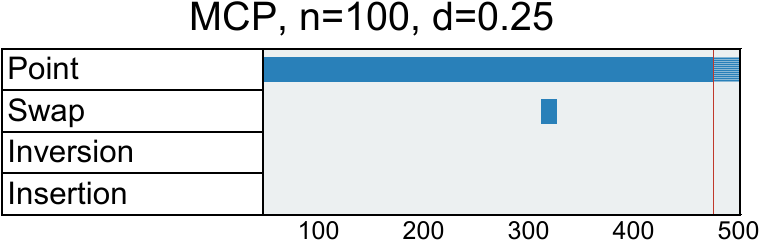}
    \end{subfigure}

    \vspace{2mm}
    
    \begin{subfigure}{0.32\textwidth}
        \includegraphics[width=\textwidth]{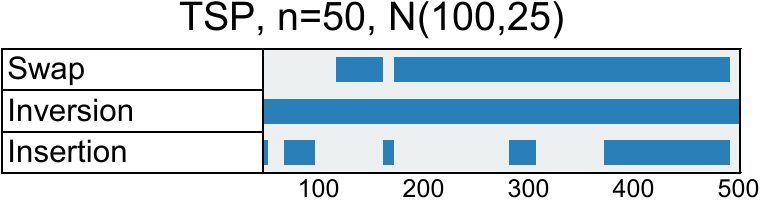}
    \end{subfigure}
    \hfill
    \begin{subfigure}{0.32\textwidth}
        \includegraphics[width=\textwidth]{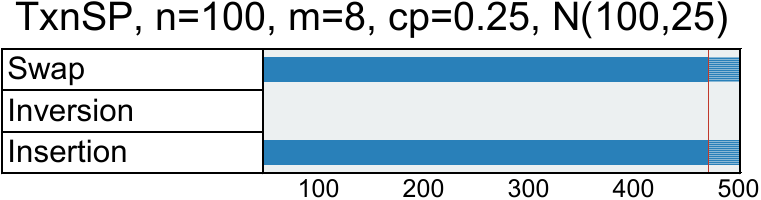}
    \end{subfigure}
    \hfill
    \begin{subfigure}{0.32\textwidth}
        \includegraphics[width=\textwidth]{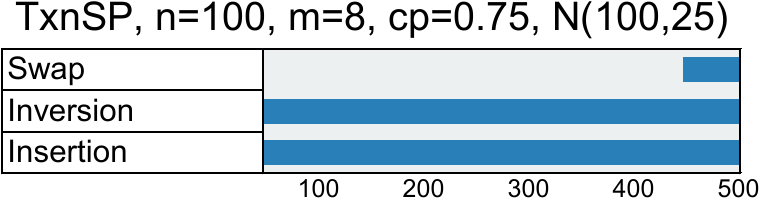}
    \end{subfigure}
    \caption{Utilization of the mutation operators by the best-performing GA configurations. The horizontal axes show the time limits in milliseconds and the blue segments mark the time limits for which each operator is selected. The vertical red line indicates the stabilization time limit.}
    \label{fig:gamutation}
\end{figure*}

    \item \textbf{Mutation Operators:} Similar to crossover, mutation operators also demonstrated the benefits of operator combinations and problem-specific preferences. For IMSP, Swap mutation was exclusively selected for all instance classes and all time limits. For small MCP instances, Swap mutation was used more frequently, while for large instances, Point was favored more. Inversion was not selected, and Insertion was used briefly. Similarly, for WMCP, Point and Swap mutations were used for all instance classes, while the other operators had limited roles. For TSP, Inversion mutation was used for the entire time limit range of all instance classes and was frequently combined with Insertion mutation. Swap mutation was only used for small instances. For TxnSP classes, the effectiveness of the mutation operators depended on $cp$. For smaller $cp$ values, Swap mutation was used more frequently. However, as $cp$ increased, Swap mutation generally became less prominent, while the frequency of Inversion mutation increased, particularly for instance classes with $n=100$. Insertion mutation was also used extensively for most TxnSP instance classes. Figure~\ref{fig:gamutation} illustrates these results.
\end{itemize}

\subsubsection{SA Operators}

\begin{itemize}[itemsep=1mm, leftmargin=0mm, label={}]

\begin{figure*}
    \begin{subfigure}{0.32\textwidth}
        \includegraphics[width=\textwidth]{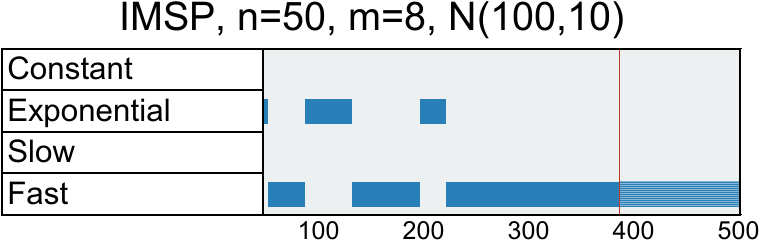}
    \end{subfigure}
    \hfill
    \begin{subfigure}{0.32\textwidth}
        \includegraphics[width=\textwidth]{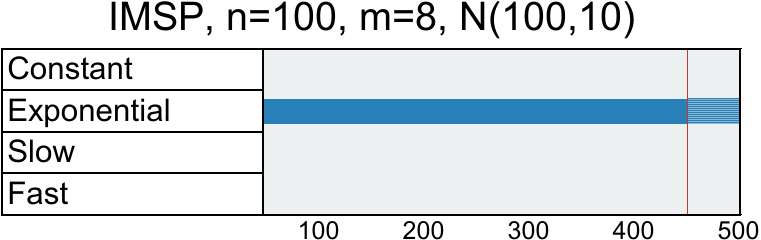}
    \end{subfigure}
    \hfill
    \begin{subfigure}{0.32\textwidth}
        \includegraphics[width=\textwidth]{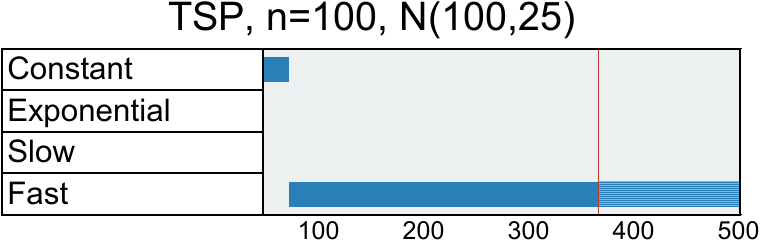}
    \end{subfigure}

    \vspace{2mm}
    
    \begin{subfigure}{0.32\textwidth}
        \includegraphics[width=\textwidth]{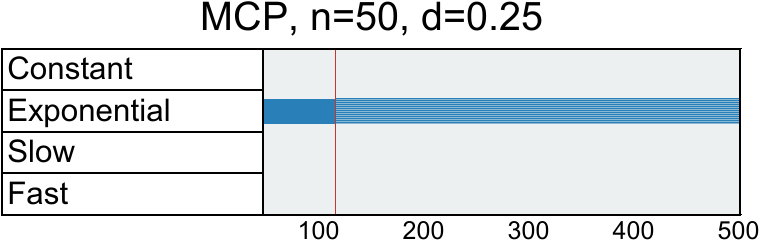}
    \end{subfigure}
    \hfill
    \begin{subfigure}{0.32\textwidth}
        \includegraphics[width=\textwidth]{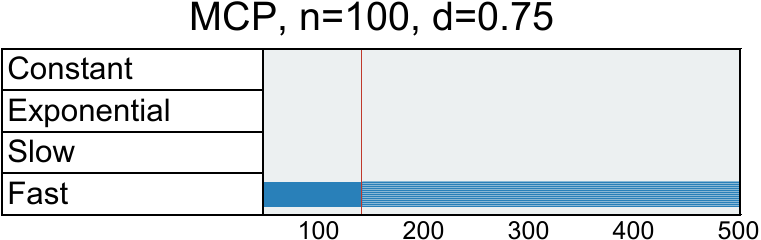}
    \end{subfigure}
    \hfill
    \begin{subfigure}{0.32\textwidth}
        \includegraphics[width=\textwidth]{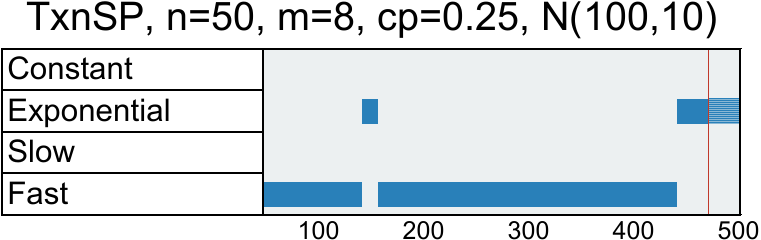}
    \end{subfigure}
    \caption{Utilization of the cooling schedules by the best-performing SA configurations. The horizontal axes show the time limits in milliseconds and the blue segments mark the time limits for which each cooling schedule is selected. The vertical red line indicates the stabilization time limit.}
    \label{fig:sacooling}
\end{figure*}

    \item \textbf{Cooling Schedules:} Fast and Exponential cooling schedules were dominant across all instance classes, whereas the Constant schedule was used briefly for only one instance class, and Slow cooling was not selected at all. For small IMSP instances ($n=50$), Fast cooling was selected much more frequently than Exponential cooling. For $n=100$, instances with $m=8$ favored Exponential cooling, while those with $m=4$ showed more mixed behavior. For two small instance classes of MCP, Exponential cooling dominated, while for the other class with $n=50$ and $d=0.50$, Fast cooling was used more frequently before stabilization. All MCP instance classes with $n=100$ favored Fast cooling. For all WMCP classes, Fast cooling dominated before stabilization. For TSP, Fast cooling was generally used, except for $n=50$ and $N(100,25)$, where Exponential cooling was also prominent before stabilization. No consistent pattern was observed for TxnSP instance classes. Figure~\ref{fig:sacooling} presents representative examples.

\begin{figure*}
    \begin{subfigure}{0.32\textwidth}
        \includegraphics[width=\textwidth]{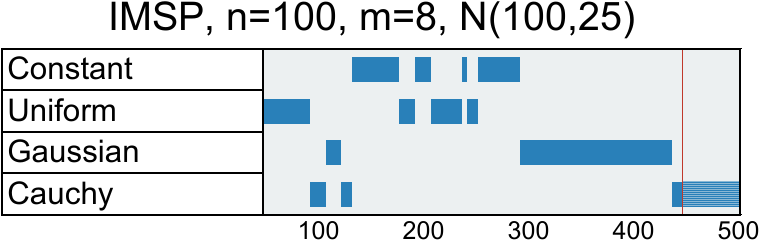}
    \end{subfigure}
    \hfill
    \begin{subfigure}{0.32\textwidth}
        \includegraphics[width=\textwidth]{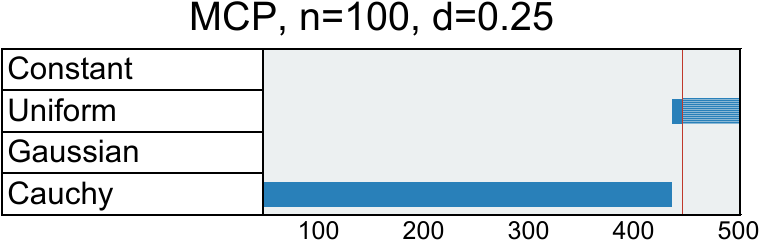}
    \end{subfigure}
    \hfill
    \begin{subfigure}{0.32\textwidth}
        \includegraphics[width=\textwidth]{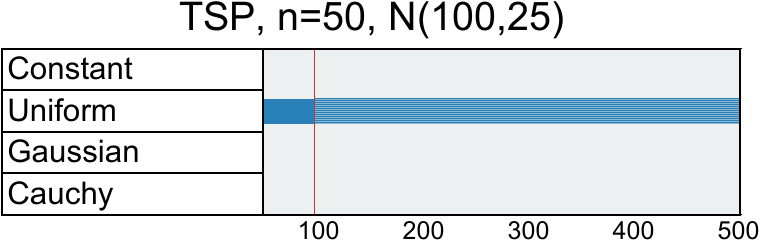}
    \end{subfigure}
    \caption{Utilization of the distance calculation operators by the best-performing SA configurations. The horizontal axes show the time limits in milliseconds and the blue segments mark the time limits for which each operator is selected. The vertical red line indicates the stabilization time limit.}
    \label{fig:sadistance}
\end{figure*}

\begin{figure*}
    \begin{subfigure}{0.32\textwidth}
        \includegraphics[width=\textwidth]{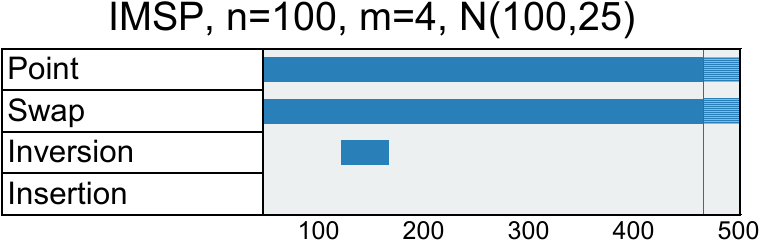}
    \end{subfigure}
    \hfill
    \begin{subfigure}{0.32\textwidth}
        \includegraphics[width=\textwidth]{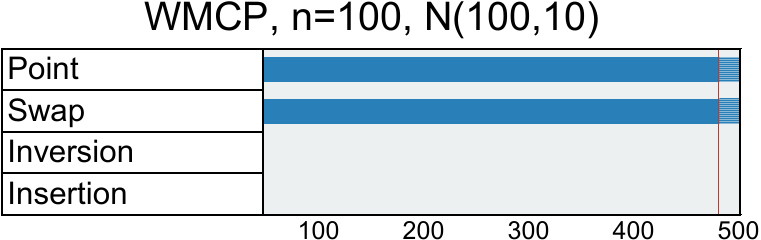}
    \end{subfigure}
    \hfill
    \begin{subfigure}{0.32\textwidth}
        \includegraphics[width=\textwidth]{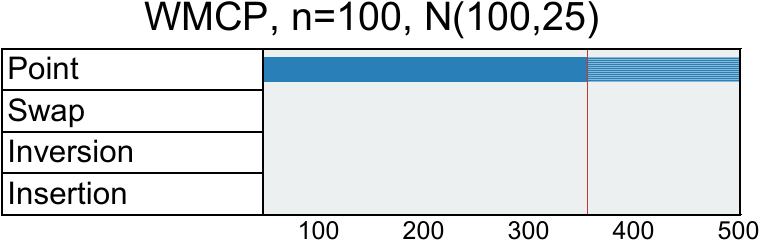}
    \end{subfigure}

    \vspace{2mm}
    
    \begin{subfigure}{0.32\textwidth}
        \includegraphics[width=\textwidth]{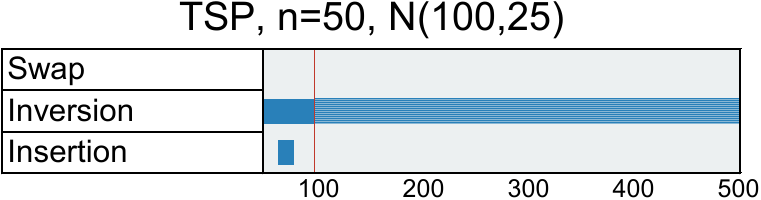}
    \end{subfigure}
    \hfill
    \begin{subfigure}{0.32\textwidth}
        \includegraphics[width=\textwidth]{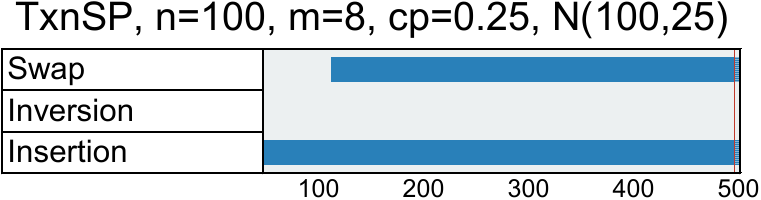}
    \end{subfigure}
    \hfill
    \begin{subfigure}{0.32\textwidth}
        \includegraphics[width=\textwidth]{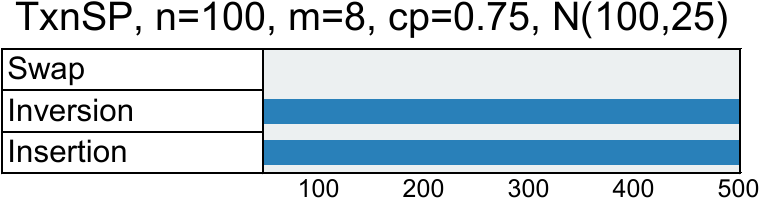}
    \end{subfigure}
    \caption{Utilization of the move operators by the best-performing SA configurations. The horizontal axes show the time limits in milliseconds and the blue segments mark the time limits for which each operator is selected. The vertical red line indicates the stabilization time limit.}
    \label{fig:samove}
\end{figure*}
    
    \item \textbf{Distance Calculation Operators:}
    Distance calculation was the most context-sensitive operator choice for SA. The proposed discrete distance calculation operators were utilized extensively by the best-performing configurations, demonstrating the usefulness of varying the neighborhood size in discrete optimization. IMSP and TxnSP presented considerable variation, and no notable pattern was found for them. Small MCP classes also showed similar behavior, as a different distance calculation operator was selected most frequently for each of them before stabilization. Cauchy distance calculation dominated large MCP classes and was the most prominent operator for all WMCP classes. For TSP classes with $n=50$, Uniform distance calculation was selected most frequently before stabilization, while no consistent pattern was observed for large instances. Figure~\ref{fig:sadistance} shows these results.

    \item \textbf{Move Operators:} Move operator selection was mostly problem-dependent, with only small differences between instance classes. TxnSP was the main exception to this generalization, as $cp$ strongly affected operator selection. Swap move was more prominent for lower $cp$ values, while as $cp$ increased, Insertion and Inversion were used much more frequently. For IMSP, Point and Swap move operators were used for the entire time limit range of every instance class, while the other move operators were occasionally used in combination with them. Similarly, Point move was used throughout every MCP and WMCP class. For small MCP classes, Swap and Insertion were also occasionally added to the combination, particularly for $d=0.25$. For WMCP instance classes with low standard deviation, Swap move was also used for the entire time limit range. For TSP, Inversion was used for all time limits and instance classes, while Insertion was additionally selected for some time limits of the high-standard-deviation instance classes. Figure~\ref{fig:samove} illustrates these findings.
\end{itemize}

\subsection{Evaluation with Benchmark Instances}

We performed additional experiments using established benchmark instances. The purpose of these experiments was to assess whether the results obtained from the synthetic analysis generalize beyond the synthetic instances and to identify the main divergences between the two sets of experiments. Similar to the previous analysis, we determined the best-performing configuration of each algorithm family for each time limit. We solved each benchmark instance multiple times with each solver configuration and used the average objective value as the measure of solution quality. 

\subsubsection{Algorithm Families}

\begin{figure*}
    \centering
    \begin{subfigure}{0.32\textwidth}
        \includegraphics[width=\textwidth]{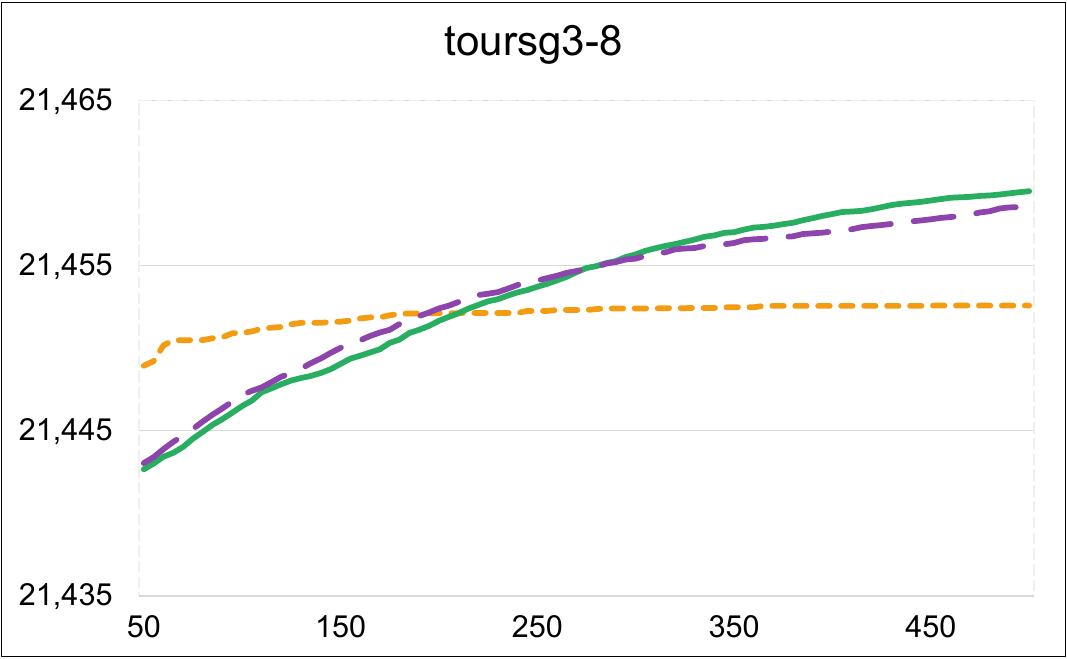}
    \end{subfigure}
    \hfill
    \begin{subfigure}{0.32\textwidth}
        \includegraphics[width=\textwidth]{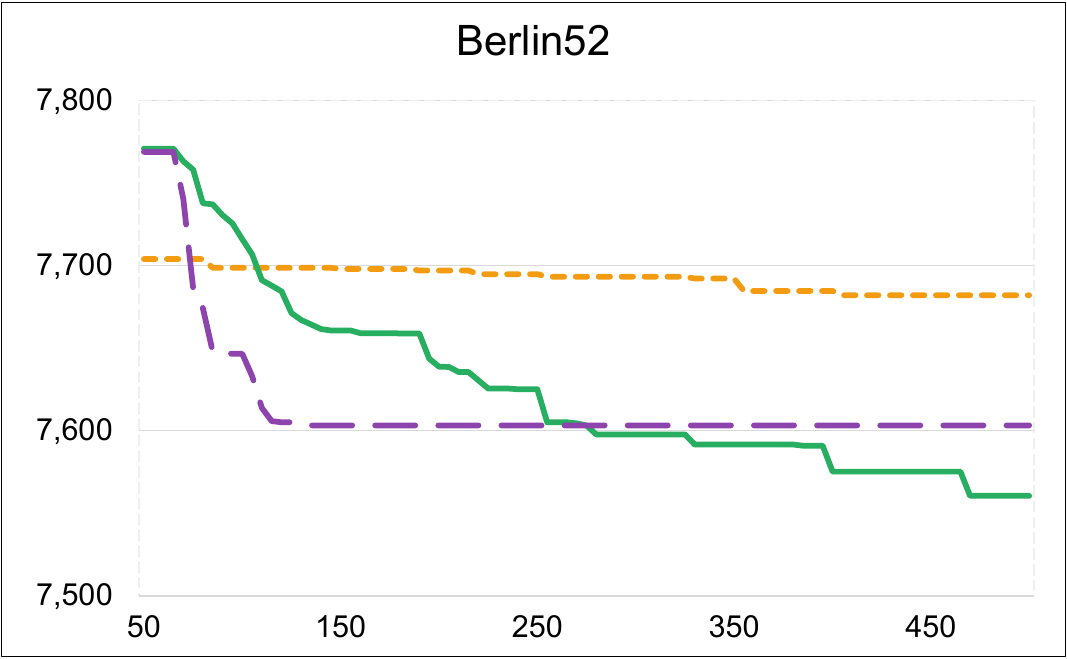}
    \end{subfigure}
    \hfill
    \begin{subfigure}{0.32\textwidth}
        \includegraphics[width=\textwidth]{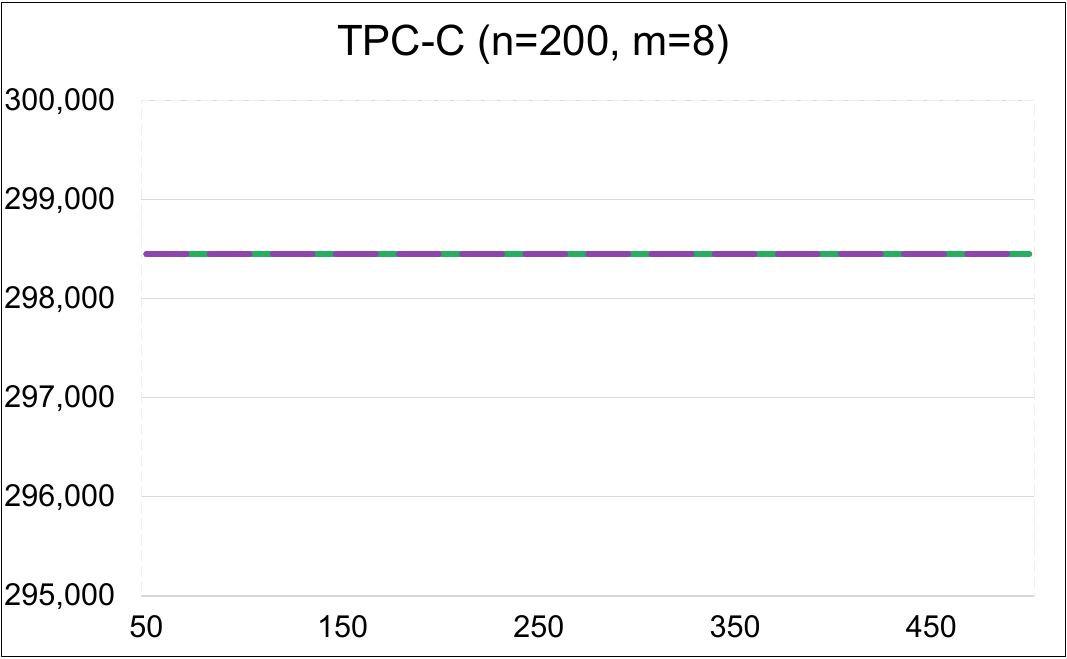}
    \end{subfigure}
    
    \caption{Comparison of the algorithm families for various time limits. The horizontal axes show the time limits in milliseconds and the vertical axes show the average objective values achieved by the best-performing configuration of each family under the associated time limit. For Berlin52 and TPC-C (n=200, m=8), a lower objective value is better, whereas for toursg3-8, a higher objective value is better. The best-performing configurations of {\color{flatorange} ACO}, {\color{flatgreen} GA}, and {\color{flatpurple} SA} are represented by {\color{flatorange} orange short-dashed}, {\color{flatgreen} green solid} and {\color{flatpurple} purple long-dashed} lines, respectively.}
    \label{fig:benchmarkgraph}
    
\end{figure*}

Figure~\ref{fig:benchmarkgraph} presents representative benchmark results illustrating the most important deviations from the synthetic-instance analysis. The results largely coincided with the results of the synthetic analysis for IMSP. GA and SA achieved similar solution qualities and both stabilized early, while ACO was not competitive for any tested instance or time limit. For MCP/WMCP, benchmark results also largely supported the synthetic analysis results. SA dominated three of the four tested instances. The only exception was toursg3-8, for which ACO achieved the best solution quality under short time limits and GA obtained the best results under long time limits. These results support the conclusion that SA is generally the strongest choice for MCP/WMCP among the tested families, while also showing that ACO and GA can be more effective for certain instances. 

For TSP, the benchmark instances were generated using positions in a two-dimensional plane, whereas synthetic instances were generated by independently sampling the distance values. This results in an inherently different instance structure, which may explain the larger deviation from the synthetic analysis results. ACO did not dominate any instance over the entire time limit range, although it achieved the best solution quality for the majority of time limits for Eil76 and under short time limits for Berlin52 and Tsp225, which indicates the impact of instance structure on algorithm selection for TSP. Another notable difference from the synthetic analysis was that GA was the most prominent algorithm family under long time limits for Berlin52, showing the potential usefulness of GA for some TSP instances.

TxnSP benchmark results showed the largest divergence from the synthetic analysis. For every tested instance, all three algorithms already achieved the same objective value at the shortest time limits and showed no further improvement throughout the tested range. This suggests that the benchmark instances were considerably easier than the synthetic instances and were unable to distinguish between the algorithm families. This behavior is consistent with the fact that the TPC-C transactions are generated from a limited set of transaction templates with fixed conflict structures and a small variation in transaction lengths, whereas in the synthetic instances, any pair of transactions can conflict and lengths are generated independently.

\subsubsection{ACO Operators}

\begin{figure*}[t]
    \centering

    \begin{subfigure}[t]{0.32\textwidth}
        \centering
        \includegraphics[width=\linewidth]{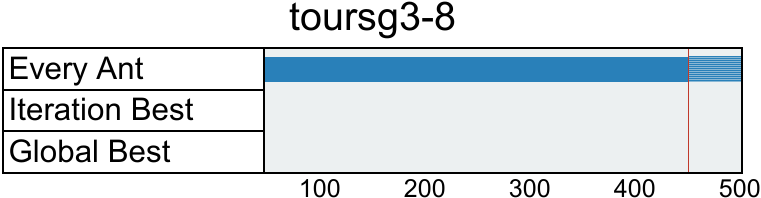}

        \vspace{2mm}

        \includegraphics[width=\linewidth]{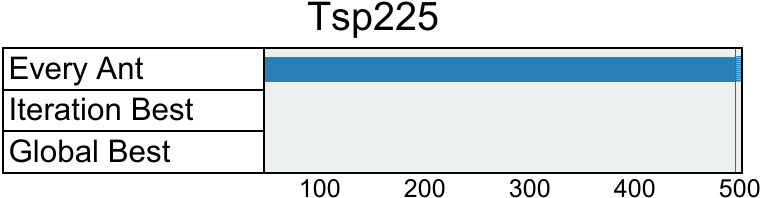}
        \caption{Tour Selection}
    \end{subfigure}
    \hfill
    \begin{subfigure}[t]{0.32\textwidth}
        \centering
        \includegraphics[width=\linewidth]{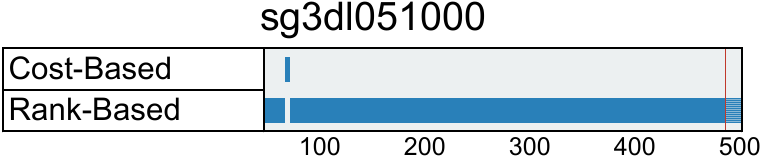}

        \vspace{2mm}

        \includegraphics[width=\linewidth]{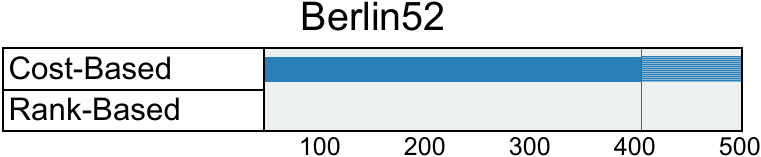}
        \caption{Pheromone Calculation}
    \end{subfigure}
    \hfill
    \begin{subfigure}[t]{0.32\textwidth}
        \centering
        \includegraphics[width=\linewidth]{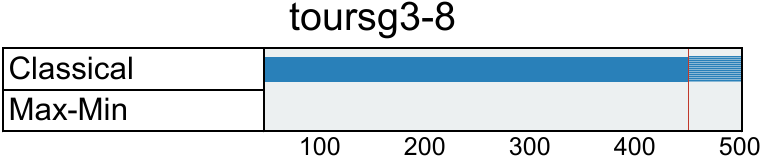}

        \vspace{2mm}

        \includegraphics[width=\linewidth]{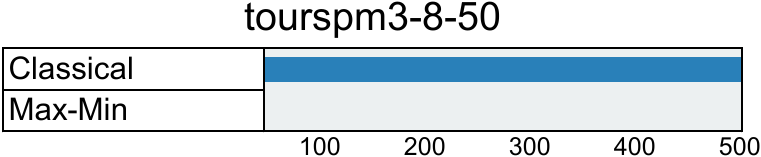}
        \caption{Pheromone Update}
    \end{subfigure}

    \caption{Representative ACO operator-utilization results for the benchmark instances. The horizontal axes show the time limits in milliseconds and the blue segments indicate the time limit ranges for which the corresponding operator is included in the best-performing configuration. The vertical red line indicates the stabilization time limit.}
    \label{fig:acobenchmark}
\end{figure*}

Since all TxnSP benchmark instances stabilized at the shortest time limit, we did not use them to draw conclusions about operator selection. Figure~\ref{fig:acobenchmark} shows representative cases for some notable benchmark results. The benchmark results showed no divergence in graph structures, as Single-Stage was used for all instances with $u=true$ while Multi-Stage structure was used for all instances with $u=false$. The results for tour selection operators also largely coincided with the synthetic analysis, with Iteration Best being the most frequently selected operator for most instances. However, Every Ant dominated the MCP/WMCP instance toursg3-8 and TSP instance Tsp225, showing that it can be useful for certain larger instances. The benchmark results supported the context dependence of pheromone calculation operators, as many tested instances alternated between the two operators. Some MCP/WMCP and TSP instances showed a stronger preference for one of these operators. Rank-Based calculation dominated sg3dl051000, sg3dl052000 and Tsp225, whereas Cost-Based calculation was utilized throughout Berlin52 and toursg3-8. The patterns observed for pheromone update operators were also largely supported by the benchmark results. Max-Min update dominated all IMSP instances and small MCP/WMCP instances (sg3dl051000 and sg3dl052000), and Classical update became more prominent for the largest TSP instance. The main divergence was observed for the two larger MCP/WMCP instances (toursg3-8 and tourspm3-8-50), for which Classical update was used throughout the entire time limit range. This suggests that the preferred pheromone update operator may change for larger or structurally different instances of these problems. 

\subsubsection{GA Operators}

\begin{figure*}[t]
    \centering

    \begin{subfigure}[t]{0.32\textwidth}
        \centering
        \includegraphics[width=\linewidth]{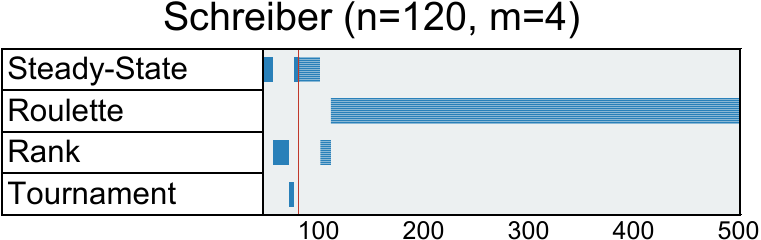}

        \vspace{2mm}

        \includegraphics[width=\linewidth]{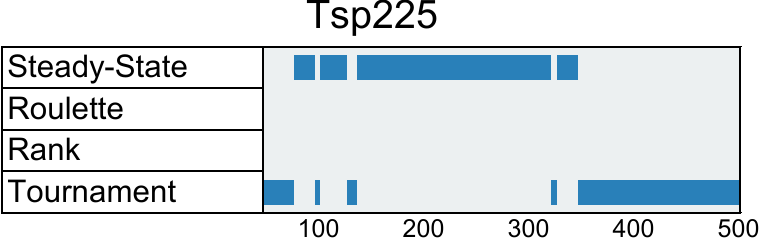}
        \caption{Selection}
    \end{subfigure}
    \hfill
    \begin{subfigure}[t]{0.32\textwidth}
        \centering
        \includegraphics[width=\linewidth]{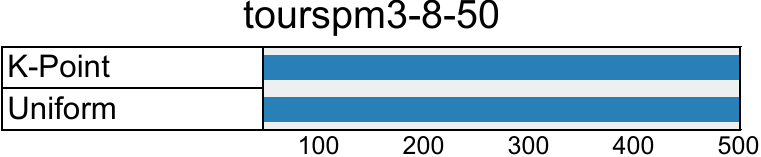}

        \vspace{2mm}

        \includegraphics[width=\linewidth]{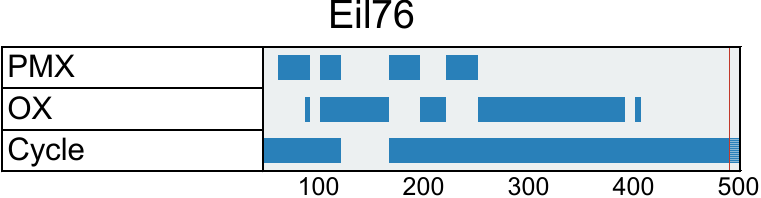}
        \caption{Crossover}
    \end{subfigure}
    \hfill
    \begin{subfigure}[t]{0.32\textwidth}
        \centering
        \includegraphics[width=\linewidth]{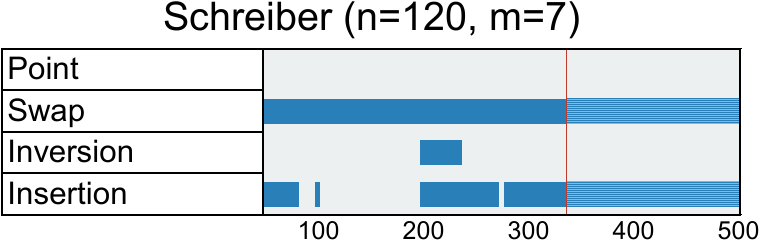}

        \vspace{2mm}

        \includegraphics[width=\linewidth]{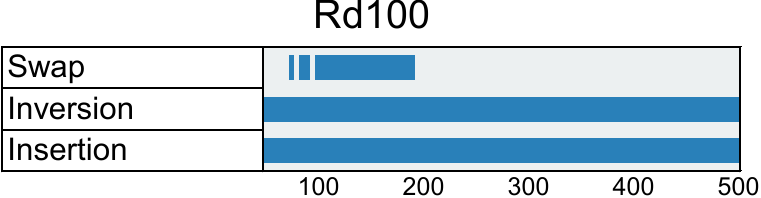}
        \caption{Mutation}
    \end{subfigure}

    \caption{Representative GA operator-utilization results for the benchmark instances. The horizontal axes show the time limits in milliseconds and the blue segments indicate the time limit ranges for which the corresponding operator is included in the best-performing configuration. The vertical red line indicates the stabilization time limit.}
    \label{fig:gabenchmark}
\end{figure*}

TxnSP instances were also not used to analyze the GA operator selection due to their very early stabilization. Some representative benchmark cases are presented in Figure~\ref{fig:gabenchmark}. IMSP instances with $m=7$ showed similar results to the synthetic analysis, as Steady-State and Tournament selection were the most prominent operators. However, the instances with $m=4$ stabilized early and did not show a clear preference before stabilization. There was considerable variation among MCP/WMCP instances, with different instances favoring different operators. Most TSP instances also showed behavior similar to the synthetic analysis, with Tournament dominating three of the four tested instances. Only Tsp225 diverged as Steady-State and Tournament were utilized with similar frequencies. The benchmark results for crossover operators largely supported the synthetic analysis. For each IMSP instance, K-Point crossover was the only operator utilized before stabilization. For MCP/WMCP, K-Point and Uniform were both used extensively, frequently in combination, with their relative utilization depending on the instance. TSP also showed behavior similar to the synthetic analysis, as Cycle was the most consistently selected crossover operator. The results for the mutation operators showed the strongest agreement with the synthetic analysis. Swap mutation dominated IMSP instances, while Point mutation was not used at all and the other operators were used only occasionally. For MCP/WMCP, Point and Swap were the most prominent mutation operators, while Insertion was selected only for brief intervals and Inversion was not selected. Inversion was used throughout all TSP instances and was frequently combined with Insertion, while Swap was used only briefly.

\subsubsection{SA Operators}

\begin{figure*}[t]
    \centering

    \begin{subfigure}[t]{0.32\textwidth}
        \centering
        \includegraphics[width=\linewidth]{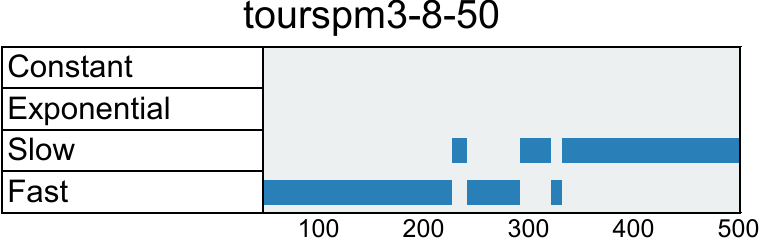}

        \vspace{2mm}

        \includegraphics[width=\linewidth]{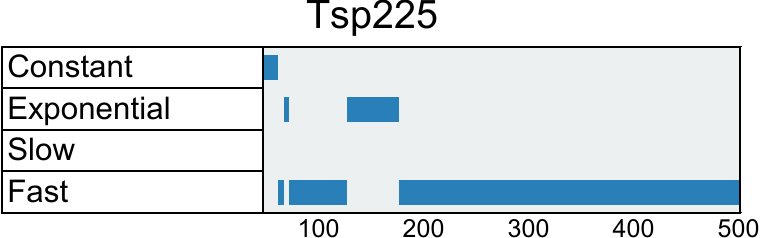}
        \caption{Cooling Schedule}
    \end{subfigure}
    \hfill
    \begin{subfigure}[t]{0.32\textwidth}
        \centering
        \includegraphics[width=\linewidth]{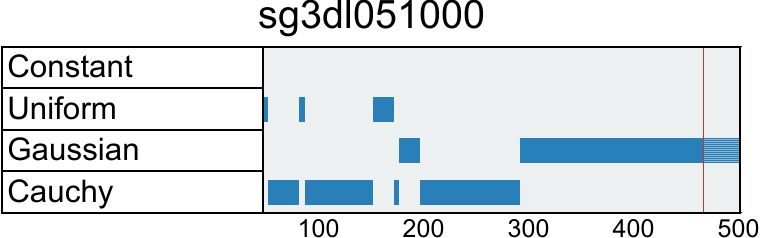}

        \vspace{2mm}

        \includegraphics[width=\linewidth]{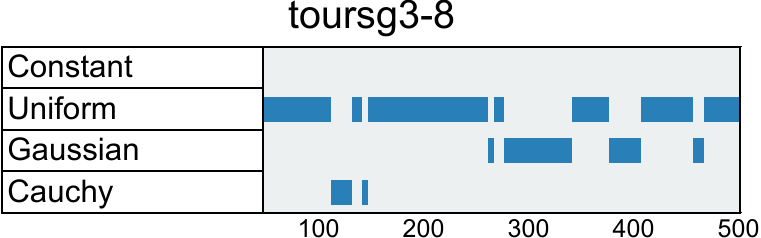}
        \caption{Distance Calculation}
    \end{subfigure}
    \hfill
    \begin{subfigure}[t]{0.32\textwidth}
        \centering
        \includegraphics[width=\linewidth]{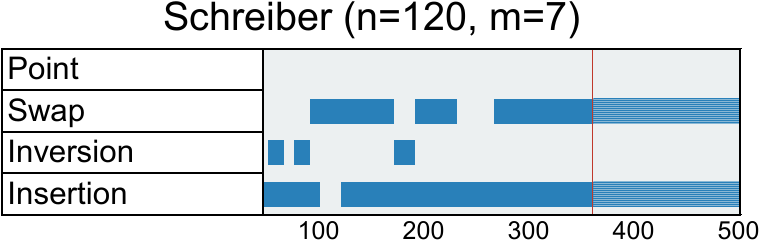}

        \vspace{2mm}

        \includegraphics[width=\linewidth]{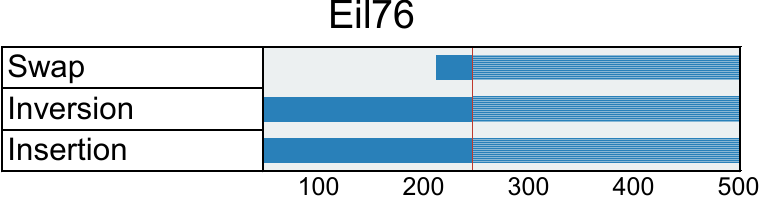}
        \caption{Move}
    \end{subfigure}

    \caption{Representative SA operator-utilization results for the benchmark instances. The horizontal axes show the time limits in milliseconds and the blue segments indicate the time limit ranges for which the corresponding operator is included in the best-performing configuration. The vertical red line indicates the stabilization time limit.}
    \label{fig:sabenchmark}
\end{figure*}

TxnSP instances were not included in the SA operator analysis, since they did not improve after 50 ms in any experiment. Figure~\ref{fig:sabenchmark} shows some representative cases from the benchmark analysis. The results for cooling schedule selection mostly agreed with the synthetic analysis, as Fast and Exponential cooling were selected most frequently. For IMSP, only Exponential cooling was used before stabilization. Fast and Exponential cooling dominated most MCP/WMCP instances, while tourspm3-8-50 differed substantially by utilizing Slow cooling over a large portion of the time limit range. TSP results also supported the synthetic analysis, as Fast and Exponential were the primary cooling schedules, Constant was used only briefly, and Slow was not selected at all. The benchmark results supported the usefulness of varying discrete neighborhood size, as Constant distance was selected briefly for only two benchmark instances before stabilization. Cauchy was the most prominent operator for IMSP. No consistent pattern was observed for MCP/WMCP, with Uniform, Gaussian, and Cauchy distance calculation becoming important for different instances. For TSP, Uniform was selected most frequently, while Gaussian and Cauchy were also utilized considerably and Constant distance was used very briefly. IMSP showed an important divergence from the synthetic analysis, as Point move was not selected before stabilization for any benchmark instance. Swap and Insertion were the most prominent operators for IMSP. MCP/WMCP results strongly supported the synthetic analysis, as Point move was used for the entire time limit range of every instance and was frequently combined with Swap. TSP also showed strong agreement, with Inversion and Insertion being utilized throughout every instance and Swap being used only occasionally.

\section{Conclusion and Future Work}
\label{sec:conclusion}

In real-time systems, optimization tasks are often subject to strict time limits. We proposed STILO, a general-purpose optimization environment for such tasks, with fine-grained configuration spaces for three well-known metaheuristic algorithm families. This framework includes novel operators and problem-independent structures, as well as combinations of operators adopted from existing variants. We also equipped STILO with analysis tools for evaluating the relative effectiveness of different algorithm families and operators under varying time limits. We identified several notable patterns using synthetic problem instances with diverse characteristics and supported these findings with benchmark instances. The results demonstrated the usefulness of the proposed discrete distance calculation mechanism for simulated annealing and showed that different problem-independent graph structures could be effective under different computational budgets. In general, our results indicate that the operators appearing in best-performing configurations depend not only on the problem type, but also on instance characteristics and the available time limit.

The STILO framework and the analytical findings of this paper provide an important basis for studying optimization under strict time limits. A natural next step is to apply the framework to real-world use cases in which discrete optimization constitutes a computational bottleneck. The framework could also be extended with additional metaheuristic algorithm families and operators. Finally, evaluating the configurations integrated into STILO on a broader range of optimization problems would provide further insight into how algorithm structures and operators should be selected for different problem settings and computational budgets.

\section*{Acknowledgments}

This work is funded by the German Federal Ministry of Research, Technology and Space within the funding program quantum technologies — from basic research to market — contract number 13N16090.


\bibliographystyle{plain}
\bibliography{references}
\end{document}